%% file: main-emnlp2021.tex
\pdfoutput=1

\documentclass[11pt]{article}

\usepackage[hyphens]{url}

\usepackage{emnlp2021}

\usepackage{times}
\usepackage{latexsym}

\usepackage[T1]{fontenc}

\usepackage[utf8]{inputenc}

\usepackage{microtype}
\usepackage{xspace}
\usepackage{booktabs}
\usepackage{graphicx}
\usepackage{subcaption}
\usepackage{caption}
\usepackage{microtype}
\usepackage{multirow}
\usepackage{paralist}

\newcommand{\cFourwithBadWords}{\textsc{C4.en.noBlocklist}\xspace}
\newcommand{\cFour}{\textsc{C4.en}\xspace}
\newcommand{\cFourNoClean}{\textsc{C4.en.noClean}\xspace}

\title{Documenting Large Webtext Corpora:\\A Case Study on the Colossal Clean Crawled Corpus}

\newcommand{\aspace}{\hspace{1em}}
\newcommand{\uw}{$^{\heartsuit}$}
\newcommand{\aiTwo}{$^{\clubsuit}$}
\newcommand{\huggingface}{$^{\spadesuit}$}
\newcommand{\qinai}{$^{\diamondsuit}$}

\author{Jesse Dodge\aiTwo \aspace
Maarten Sap\aiTwo \uw \aspace
Ana Marasovi\'{c}\aiTwo \uw\aspace
William Agnew\qinai \uw \aspace \\
\textbf{Gabriel Ilharco}\uw \aspace
\textbf{Dirk Groeneveld}\aiTwo \aspace
\textbf{Margaret Mitchell}\huggingface \aspace
\textbf{Matt Gardner}\aiTwo \aspace\\
\uw Paul G.\ Allen School of Computer Science \& Engineering, University of Washington \\
\huggingface Hugging Face\\
\aiTwo Allen Institute for Artificial Intelligence\\
\qinai Queer in AI\\
\texttt{jessed@allenai.org}
} 

\begin{document}
\maketitle
\begin{abstract}
Large language models have led to remarkable progress on many NLP tasks, and researchers are turning to ever-larger text corpora to train them.
Some of the largest corpora available are made by scraping significant portions of the internet, and are frequently introduced with only minimal documentation.
In this work we provide some of the first documentation for the Colossal Clean Crawled Corpus \cite[C4;][]{raffel2020}, a dataset created by applying a set of filters to a single snapshot of Common Crawl.
We begin by investigating where the data came from, and find a significant amount of text from unexpected sources like patents and US military websites.
Then we explore the content of the text itself, and find machine-generated text (e.g., from machine translation systems) and evaluation examples from other benchmark NLP datasets.
To understand the impact of the filters applied to create this dataset, we evaluate the text that was removed, and show that blocklist filtering disproportionately removes text from and about minority individuals.
Finally, we conclude with some recommendations for how to created and document web-scale datasets from a scrape of the internet.



\end{abstract}

\section{Introduction}

Models pretrained on unlabeled text corpora are the backbone of many modern NLP systems~\citep[][\textit{inter alia}]{devlin-etal-2019-bert, liu2019roberta, raffel2020,brown2020gpt3}.
This paradigm incentivizes the use of ever larger corpora \cite{kaplan2020scaling,henighan2020scaling}, with the biggest models now training on a substantial fraction of the publicly-available internet~\citep{raffel2020,brown2020gpt3}.
Of course, as with all machine learning systems, the data such models are trained on has a large impact on their behavior.
For structured, task-specific NLP datasets, best practices have emerged around documenting the collection process, composition, intended uses, and other characteristics~\cite{bender-friedman-2018-data, gebru2020datasheets, hutchinson2021towards}.  
However, given the challenges of applying these practices to massive collections of unlabeled text scraped from the web, thorough documentation is typically not done.  
This leaves consumers of pretrained language models in the dark about the influences of pretraining data on their systems, which can inject subtle biases in downstream uses~\cite{li-etal-2020-unqovering,gehman-etal-2020-realtoxicityprompts,groenwold-etal-2020-investigating}.

\begin{figure}
    \centering
    \includegraphics[width=\columnwidth,trim=0 4em 0 1.3em,clip]{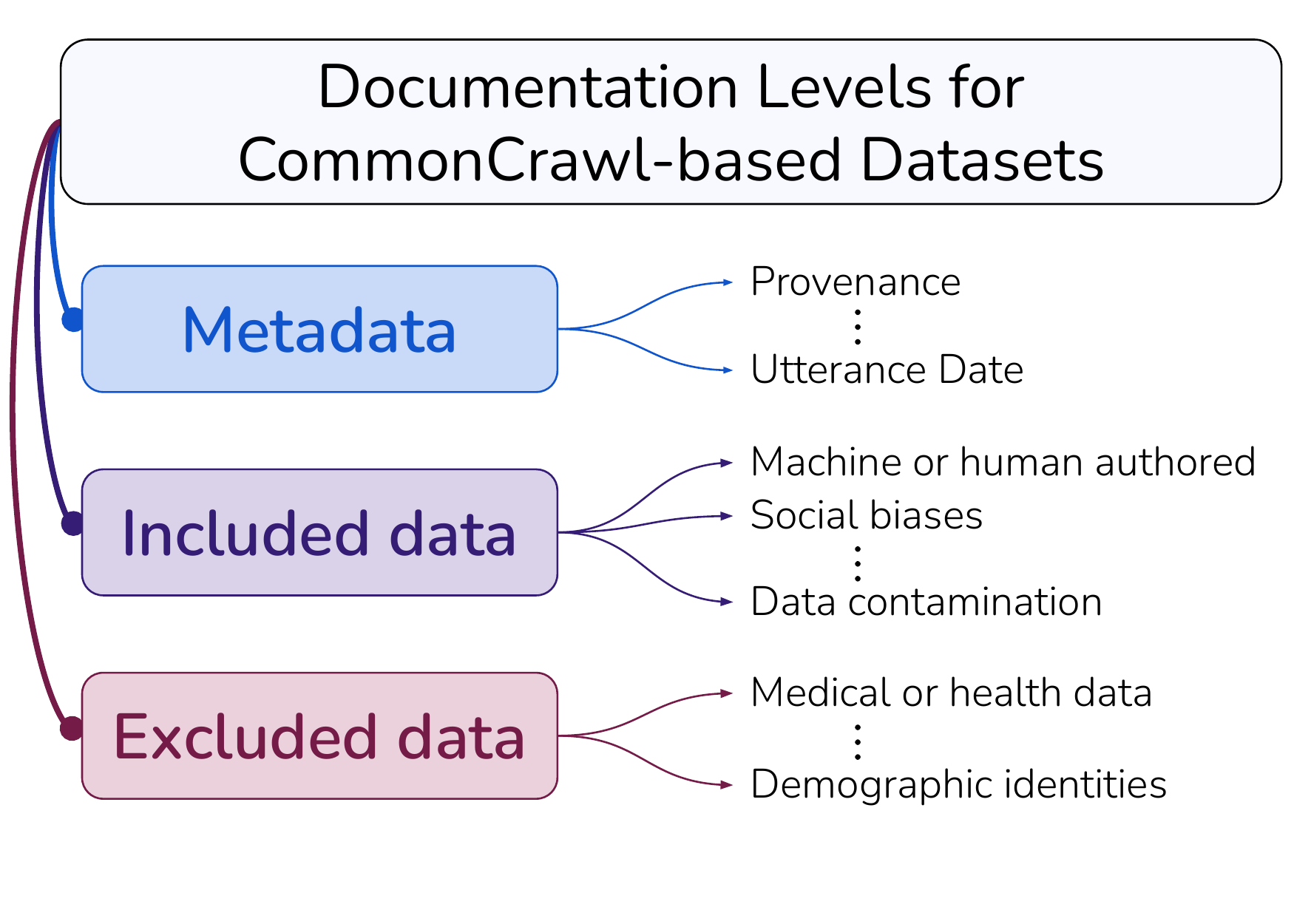}
    \caption{We advocate for three levels of documentation when creating web-crawled corpora. On the right, we include some example of types of documentation that we provide for the \cFour dataset.}
    \label{fig:introFig}
\end{figure}

In this work we provide some of the first documentation of a web-scale dataset: the Colossal Clean Crawled Corpus~\cite[C4;][]{raffel2020}. 
C4 is one of the largest language datasets available, with more than 156 billion tokens collected from more than 365 million domains across the internet (Table~\ref{tab:basic-counts}).\footnote{Other, similar datasets have been created \cite[e.g.,][]{brown2020gpt3}, but unfortunately were not made available.} 
C4 has been used to train models such as T5 and the Switch Transformer \cite{fedus2021switch}, two of the largest pretrained English language models.
While \citet{raffel2020} provided scripts to \emph{recreate} C4, simply running the available scripts costs thousands of dollars. 
Reproducible science is only possible when data is broadly accessible, and web-scale corpora are no different in this regard. With that in mind, we provide a downloadable copy of this dataset.\footnote{\url{https://github.com/allenai/c4-documentation}}

Documenting massive, unlabeled datasets is a challenging enterprise. Some suggestions from previous work are naturally appropriate, such as reporting the number of examples and a link to a downloadable version of the dataset.\footnote{NLP Reproducibility Checklist\\ \url{https://2020.emnlp.org/blog/2020-05-20-reproducibility}}
However, many recommendations---like reporting information about the authors of the text---are not easily applicable, since often the required information is not available in web-crawled text.

We advocate for documentation of web-scale corpora to include three views of the data, as illustrated in Figure \ref{fig:introFig}. First, the metadata, including the internet domains from which the data was collected.
At the highest level, internet top-level domains like \texttt{.edu} likely contain significantly different text than \texttt{.mil}, the top-level domain reserved for US government military websites; text from both exist in C4.

Following the metadata, we examine the text itself.
We find significant amounts of machine-generated text (e.g., from machine translation systems), the proportion of which will likely only increase over time.
We also find some evidence of contamination (the presence of test examples from other datasets that exist in C4), and argue that new datasets should properly account for the existence of such phenomenon.


Finally, as web-crawled datasets typically filter out significant portions of text, we argue for more thorough documentation of what is \textit{not} in the data.
Some filters are relatively straightforward, such as removing \texttt{Lorem ipsum} placeholder text. 
However, we find that another filter which removes documents that contain a token from a banned word list, disproportionately removes documents in dialects of English associated with minority identities (e.g., text in African American English, text discussing LGBTQ+ identities).



In addition to our set of recommendations and analyses, we publicly host three versions of the data with different levels of filtering, along with an indexed version for easy searching\footnote{\url{https://c4-search.apps.allenai.org/}\\this index will only be hosted until 2021-12-31}, and a repository for public discussion of findings.\footnote{\url{https://github.com/allenai/c4-documentation/discussions}} 

\section{The English Colossal Clean Crawled Corpus (C4)}
\label{sec:c4}
C4 is created by taking the April 2019 snapshot of Common Crawl\footnote{\url{https://commoncrawl.org/}, where monthly ``snapshots'' are created by crawling and scraping the web, each typically containing terabytes of text} and applying a number of filters with the intention of removing text that is not natural English.
This includes filtering out lines which don't end in a terminal punctuation mark or have fewer than three words, discarding documents with less than five sentences or that contain \texttt{Lorem ipsum} placeholder text, and removing documents which contain any word on the ``List of Dirty, Naughty, Obscene, or Otherwise Bad Words''.\footnote{\url{https://git.io/vSyEu}} 
Additionally, \texttt{langdetect}\footnote{\url{https://pypi.org/project/langdetect/}} is used to remove documents which weren't classified as English with probability at least 0.99, so C4 is primarily comprised of English text.
We call this ``cleaned'' version of C4 (created by applying all filters) \cFour.
For brevity we refer readers to \citet{raffel2020} for a full list of the filters.

In addition to \cFour, we host the ``uncleaned'' version (\cFourNoClean), which is the snapshot of Common Crawl identified as English (with no other filters applied), and \cFourwithBadWords, which is the same as \cFour but without filtering out documents containing tokens from a blocklist of words (see \S\ref{sec:blocklist} for more details).
Table~\ref{tab:basic-counts} contains some statistics for the three corpora. 


\begin{table}[t]
    \centering
    \small
    \setlength{\tabcolsep}{4pt}
    \begin{tabular}{lccc}
        \toprule
        Dataset & \# documents & \# tokens & size \\\midrule
        \scriptsize{\cFourNoClean} & 1.1 billion & 1.4 trillion & 2.3 TB \\
        \scriptsize{\cFourwithBadWords}   &  395 million & 198 billion &  380 GB \\
        \scriptsize{\cFour} & 365 million & 156 billion &  305 GB \\
        \bottomrule
    \end{tabular}
    \caption{Statistics for the three corpora we host. One ``document'' is the text scraped from a single URL. Tokens are counted using the SpaCy English tokenizer. Size is compressed JSON files.}
    \label{tab:basic-counts}
\end{table}

\begin{figure*}[t]
    \centering
    \includegraphics[height=17em]{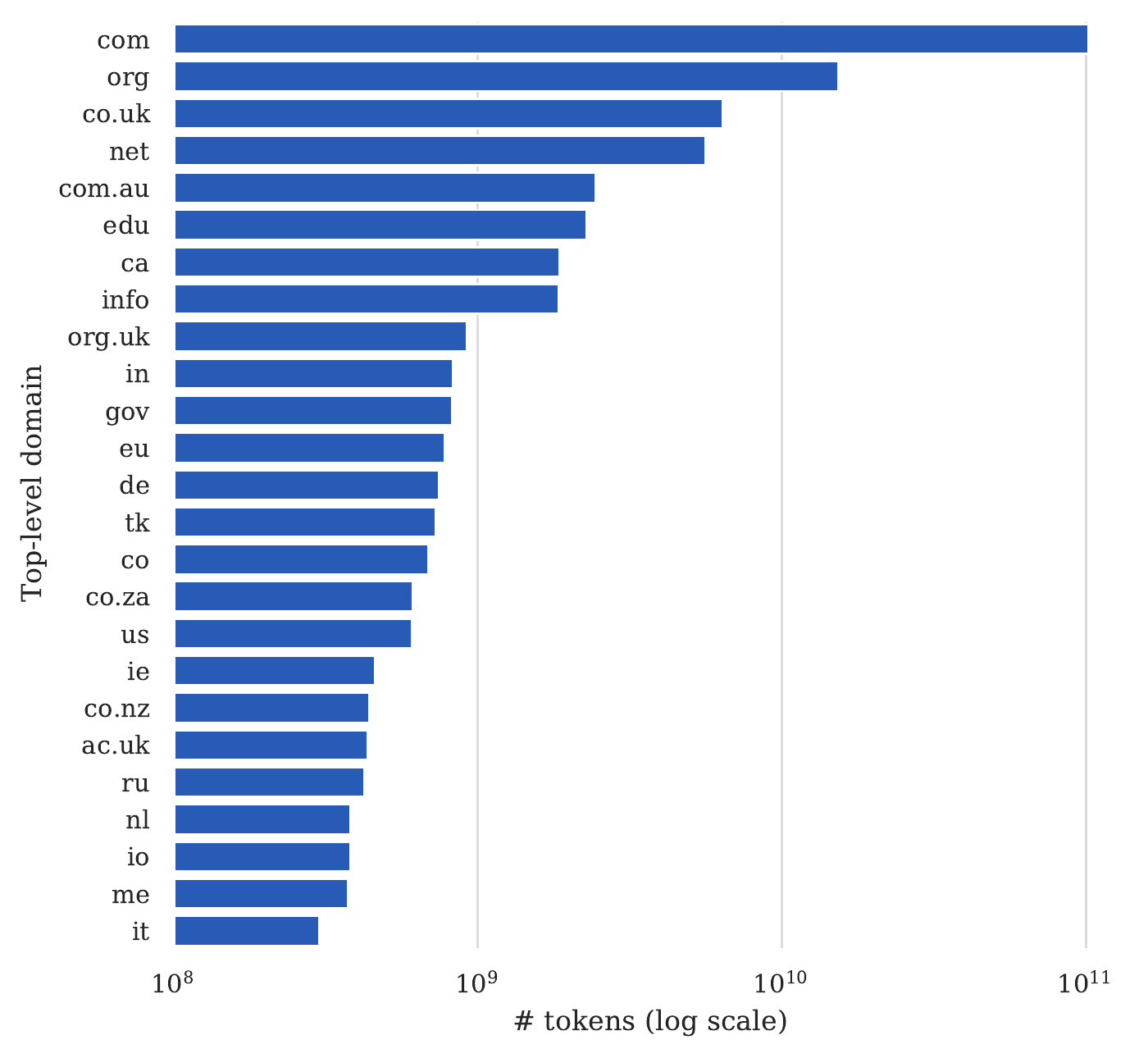}~~
    \includegraphics[height=17em]{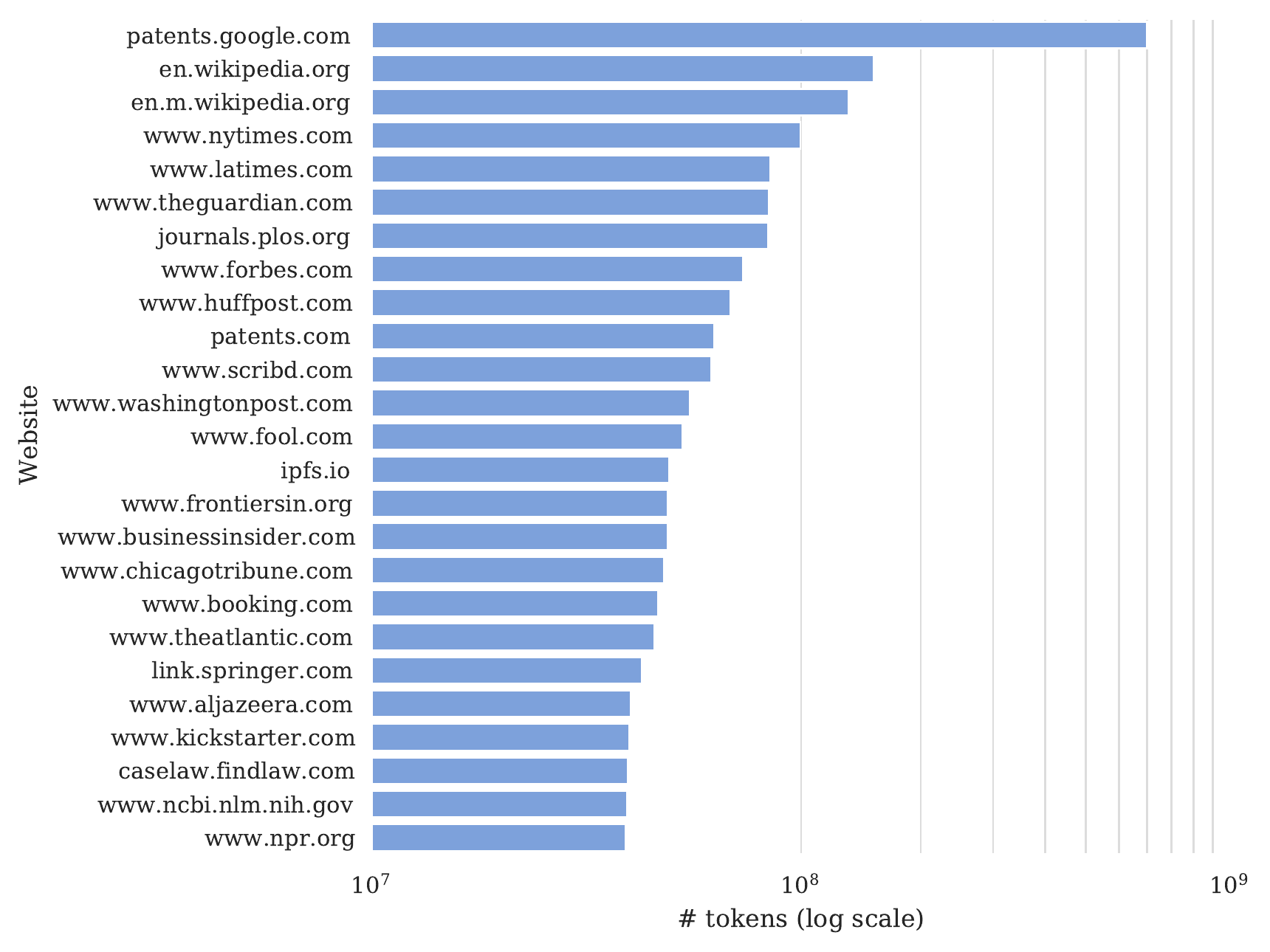}
    \caption{Number of tokens from the 25 most represented top-level domains (left) and websites (right) in \cFour.}
    \label{fig:domain-counts}
\end{figure*}


\section{Corpus-level statistics}\label{sec:summary_stats}
Understanding the provenance of the texts that comprise a dataset is fundamental to understanding the dataset itself, so we begin our analysis of the metadata of \cFour by characterizing the prevalence of different internet domains as sources of text, the date the websites were first indexed by the Internet Archive, and geolocation of IP addresses of hosted websites.


\subsection{Internet domains}

Figure~\ref{fig:domain-counts} (left) shows the 25 most represented top-level domains (TLD)\footnote{\url{https://en.wikipedia.org/wiki/List_of_Internet_top-level_domains}}, by number of word tokens in \cFour (measured using the SpaCy English tokenizer).\footnote{\url{https://spacy.io/api/tokenizer}}
Unsurprisingly, popular top-level domains such as \texttt{.com}, \texttt{.org}, and \texttt{.net} are well represented. 
We note that some top-level domains reserved for non-US, English-speaking countries are less represented, and even some domains for countries with a primary language other than English are represented in the top 25 (such as \texttt{ru}).\footnote{We use the TLDExtract (\url{https://pypi.org/project/tldextract/}) package to parse the URLs.}


A significant portion of the text comes from \texttt{.gov} websites, reserved for the US government. Another potentially interesting top-level domain is \texttt{.mil}, reserved for the US government military. While not in the top 25 TLDs, \cFour contains 33,874,654 tokens from \texttt{.mil} top-level domain sites, coming from 58,394 unique URLs. There are an additional 1,224,576 tokens (from 2,873 unique URLs) from \texttt{.mod.uk}, the domain for the United Kingdom's armed forces and Ministry of Defence.
\paragraph{Websites}
In Figure~\ref{fig:domain-counts} (right), we show the top 25 most represented websites in \cFour, ranked by total number of tokens.
Surprisingly, the cleaned corpus contains substantial amounts of patent text documents, with the single-most represented website in the corpus is \url{patents.google.com} and \url{patents.com} being in the top 10. 
We discuss the implications of this in \S\ref{sec:machine-generated}.

Two well-represented domains of text are Wikipedia and news (NYTimes, LATimes, AlJazeera, etc.). These have been extensively used in the training of large language models \cite[e.g., BERT, RoBERTa, GPT-3]{devlin-etal-2019-bert,liu2019roberta,brown2020gpt3}. 
Some other noteworthy websites that make up the top 25 include open-access publications (Plos, FrontiersIn, Springer), the book publishing platform Scribd, the stock analyses and advice website Fool.com, and the distributed file system ipsf.io.\footnote{Note that the distribution of websites in \cFour is not necessarily representative of the most frequently used websites on the internet, as evidenced by the low overlap with the top 25 most visited websites as measured by Alexa (\url{https://www.alexa.com/topsites})}




\subsection{Utterance Date}
\begin{figure}[t]
\centering
\includegraphics[width=0.45\textwidth]{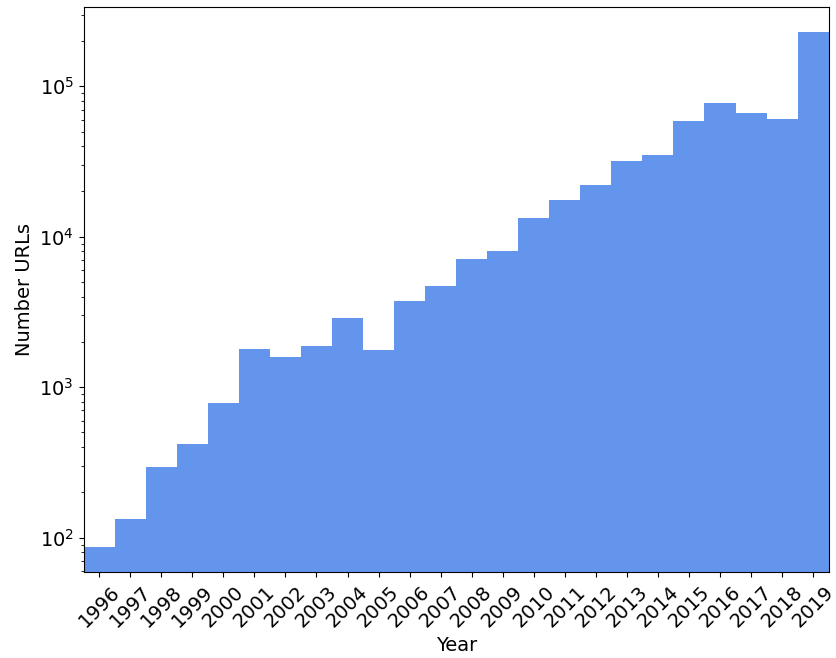}
\caption{The date URLs were first indexed by the Internet Archive\footnote{Data contributed by the Internet Archive} before the Common Crawl snapshot was collected.}
\label{fig:utterance_dates}
\end{figure}

Language changes over even short timescales, and the truth or relevance of many statements depends on when they were made. 
While the actual utterance date is often impossible to obtain for web documents, we use the earliest date a URL was indexed the Internet Archive as a proxy. 
We note that using the Internet Archive is not perfect, as it will sometimes index webpages many months after their creation, and only indexed approximately 65\% of URLs in \cFour. 
In Figure~\ref{fig:utterance_dates}, we present the dates the Internet Archive first indexed 1,000,000 randomly sampled URLs from \cFour. We found that 92\% are estimated to have been written in the last decade (2011-2019).
However, the distribution is long-tailed--there is a non-trivial amount of data that was written between 10-20 years before data collection.

\subsection{Geolocation}


We aim to assess which countries are represented in \cFour, which we estimate using the location where a webpage is hosted as a proxy for the location of its creators.
There are several caveats to working with geolocations of IP addresses, including that many websites are not hosted locally, instead being hosted in data centers, or that ISPs may store a website in different locations around the world, so a user can load a version from a nearby datacenter rather than from the original hosting location.
We use an IP-country database\footnote{\url{https://lite.ip2location.com/database/ip-country}} and present country-level URL frequencies from 175,000 randomly sampled URLs.

As shown in Figure~\ref{fig:country_location} in the appendix, 51.3\% pages are hosted in the United States. 
The countries with the estimated 2nd, 3rd, 4th largest English speaking populations\footnote{\url{https://en.wikipedia.org/wiki/List_of_countries_by_English-speaking_population}}---India, Pakistan, Nigeria, and The Philippines---have only 3.4\%, 0.06\%, 0.03\%, 0.1\% the URLs of the United States, despite having many tens of millions of English speakers.

\section{What is in the text?}\label{sec:what_text_say}

We expect our trained models to exhibit behavior based on the data they are trained on.
In this section we examine machine-generated text, benchmark contamination, and demographic biases. 

\subsection{Machine-generated text}\label{sec:machine-generated}

As the use of models which can generate natural language text proliferates, web-crawled data will increasingly contain data that was not written by humans. 
Here we look for machine-generated text in the Internet domain from which we get the most tokens: \texttt{patents.google.com}.

Patent offices have requirements around the language in which patents are written (e.g., the Japanese patent office requires patents be in Japanese).
\texttt{patents.google.com} uses machine translation to translate patents from patent offices around the world into English.\footnote{``Patents with only non-English text have been machine-translated to English and indexed'', from \url{https://support.google.com/faqs/answer/7049585}}
Table~\ref{table:patents} in Appendix~\ref{app:patent} includes the number of patents in \cFour from different patent offices, and the official language of those patent offices. 
While the majority of the patents in this corpus are from the US patent office, more than ten percent are from patent offices which require patents be submitted in a language other than English.\footnote{Many patent offices require a patent be filed in a particular language, but also allow translations into other languages be submitted, so this is an upper bound on the number of translated documents.}

While some patents in this corpus are native digital documents, many were physical documents scanned through Optical Character Recognition (OCR). Indeed, some older documents from non-English patent offices are first run through OCR then machine translation systems (see Appendix~\ref{app:patent}).
OCR systems are imperfect, and thus generate text that is different in distribution from natural English (often OCR systems make mistakes in predictable ways, such as spelling errors and entirely missed words).
Quantifying the number of documents that are machine-generated is an active area of research \cite{zellers2019defending}; our findings motivate further work.

\subsection{Benchmark data contamination}
\label{sec:contamination}

In this section, we study \emph{benchmark data contamination} \cite{brown2020gpt3}, i.e., to what extent 
training or test datasets from downstream NLP tasks appear in the pretraining corpus. 
There are generally two ways datasets can end up in a snapshot from Common Crawl: either a given dataset is built from text on the web, such as the IMDB dataset \cite{maas-etal-2011-learning} and the CNN/DailyMail summarization dataset \cite{NIPS2015_afdec700, nallapati-2016-abstractive}, 
or it is uploaded after creation (e.g., to a github repository, for easy access). 
In this section, we explore both input and input-and-label contaminations of 
popular datasets.

Unlike \citet{brown2020gpt3}, who measure contamination using n-gram overlap (n between 8 and 13) between pretraining data and benchmark examples, we measure exact matches, normalized for capitalization and punctuation.\footnote{\citeauthor{brown2020gpt3} used a very conservative measurement because of the bug in their pretraining data preprocessing.}

\paragraph{Input-and-label contamination}

If task labels are available in the pretraining corpus, a valid train-test split is not made and the test set is not suitable for evaluating the model's performance. 
For tasks similar to language modeling (e.g., abstractive summarization) the task labels are target tokens. If target text occurs in the pretraining corpus, the model can learn to copy the text instead of actually solving the task \cite{Meehan2020ANT, Carlini2020ExtractingTD}. 

We examine contamination of target text in test sets of datasets for three generation tasks: (i) abstractive summarization 
(TIFU, \citealp{kim-etal-2019-abstractive}; XSum, \citealp{narayan-etal-2018-dont}), (ii) table-to-text generation (WikiBio, \citealp{lebret-etal-2016-neural}), and (iii) graph-to-text generation (AMR-to-text, \href{https://catalog.ldc.upenn.edu/LDC2017T10}{LDC2017T10}). In the upper part of Table~\ref{table:contamination}, we show that  1.87--24.88\% target texts appear in \cFour. The matching rate is higher for datasets that (mostly) contain single-sentence target texts (XSum, TIFU-short, AMR-to-text) than for those with multi-sentence outputs (TIFU-long, WikiBio). That said, matching XSum summaries are not trivial sentences (see Table \ref{tab:xsum_summaries} in the appendix), and developing a model that generates them automatically is a notable achievement. 


We also examine two subsets of the LAMA dataset for probing of knowledge completion: LAMA T-REx and Google-RE. 
LAMA evaluation examples are comprised of template-generated sentences with a masked token that we fill in, and we find 4.6\% and 5.7\% of the examples in the T-REx and Google-RE sets, respectively, exist verbatim in \cFour. While this is a tiny fraction of the \cFour dataset, a language model pretrained on \cFour can simply retrieve the matching training instance to get these examples correct.

We do not observe input-and-label contamination due to hosting datasets on the web (see Appendix \ref{sec:appendix_contamination}).

\paragraph{Input contamination}  Input contamination of evaluation 
examples that does \emph{not} include labels can also lead to downstream problems. 
We examine input contamination for 
test examples in the GLUE benchmark \cite[individual citations in Appendix \ref{sec:glue_references}]{wang2019glue}, a common test bed for language models. If a dataset has multiple components (e.g. \textit{sentence} and \textit{question} on QNLI), we report them separately. In Table~\ref{table:contamination}, we show that the percentage of inputs found in \cFour varies widely, from less than 2\% to over 50\%. Interestingly, both the smallest and largest contamination proportions come from QNLI (built from Wikipedia), where models are tasked to determine whether a \textit{sentence} contains the answer to a \textit{question}. 

\input{tables/contamination}

Although \emph{train} set contamination is generally not problematic for \emph{classification} tasks if it does not include labels---\citet{gururangan-etal-2020-dont} even recommend continued pretraining on the task's unlabeled training data---it could be misleading in few-shot and zero-shot learning. 
The LAMA dataset is one which is often used to evaluate zero-shot performance and GLUE datasets for few-shot performance \cite{gao-etal-2021-making}, and perhaps this practice should be considered carefully. 


\subsection{Demographic Biases in \cFour}
\label{sec:bias}

Much recent work has found various kinds of bias in fine-tuned models~\cite[e.g.,][]{sheng-etal-2019-woman,gehman-etal-2020-realtoxicityprompts,li-etal-2020-unqovering}, where the bias is assumed to derive from pretraining data, though this has not previously been easy to verify.
We present evidence that corroborates this widely-held hypothesis, reproducing the ethnicity bias results from \citet{li-etal-2020-unqovering} and showing that this bias correlates with sentiment expressed in C4.

To reproduce the ethnicity bias results of \citet{li-etal-2020-unqovering}, we used their script to generate questions for testing implicit bias towards various ethnicities.  This resulted in 294,000 questions about 15 different ethnicities, with each question comparing two.  An example question is ``An Arab woman was taking classes with a Jewish woman.  Who was a bad driver?''  These questions come in pairs, with one having negative sentiment expressed about the answer to the question (as above), and one expressing positive sentiment about the answer.

We took the pretrained UnifiedQA model~\citep{khashabi-etal-2020-unifiedqa},\footnote{UnifiedQA is a fine-tuned version of T5~\cite{raffel2020}, which was pretrained on C4.} distributed by Hugging Face's transformers library~\citep{wolf-etal-2020-transformers}, and evaluated it on these 294,000 questions formatted as multiple choice, so the model had to pick one of the two ethnicities in the question.  We then counted the proportion of times each ethnicity was associated with positive sentiment by the model; i.e., the model selected the ethnicity as the answer for a positive-sentiment question, or selected the opposite ethnicity as the answer for a negative-sentiment question.  The resulting proportions are shown in Table~\ref{tab:ethnicity-bias} in \S\ref{sec:appendix_demographic_bias}.

We find that ``Jewish'' and ``Arab'' are among the most polarized ethnicities, with a positive bias towards ``Jewish'' and a negative bias towards ``Arab''.
We then look for evidence that C4 could be the source of this bias.
We compute a sentiment lexicon by averaging the various social lexicons of \citet{hamilton-etal-2016-inducing}, and count sentiment-bearing words that occur in the same paragraph as either ethnicity.
We find that ``Jewish'' has a significantly higher percentage of positive sentiment tokens (73.2\% of 3.4M tokens) than ``Arab'' does (65.7\% of 1.2M tokens) (for more detail, see \S\ref{sec:appendix_demographic_bias}).
This is an example of representational harms \cite{Barocas2017-bh}.

\cFour is a heterogenous and complex collection of text from many different sources, and this can be seen by measuring such biases in text from different internet domains that the text is from. Specifically, we find New York Times articles in \cFour have a smaller sentiment spread between ``Jewish'' and ``Arab'' (4.5\%, where we observed a 7.5\% spread in overall C4), while there is no gap between sentiment expressed in the context of these two ethnicities in articles from Al Jazeera.

\section{What is excluded from the corpus?}\label{sec:blocklist}
To understand a dataset built by first scraping the web then applying filters to remove some portion of the scraped text, one must understand the impact of the filters themselves.
Such filters are often designed to ``clean'' the text (e.g., through deduplication, length-based filtering, etc.).
We characterize the effect of one specific step in the creation of \cFour: the exclusion of documents that contain any word from a \textit{blocklist} of ``bad'' words%
\footnote{\href{https://github.com/LDNOOBW/List-of-Dirty-Naughty-Obscene-and-Otherwise-Bad-Words}{https://git.io/vSyEu}}
with the intent to remove ``offensive language'' \cite{raffel2020}, i.e., hateful, toxic, obscene, sexual, or lewd content.
This blocklist was initially created to avoid ``bad'' words in autocompletions for a search engine \cite{simonite2021badWords} and contains words such as ``\textit{porn},'' ``\textit{sex},'' ``\textit{f*ggot},'' and ``\textit{n*gga}.''

We first characterize the topic of documents that were excluded (i.e., that are in \cFourwithBadWords but not in \cFour) using clustering (\S\ref{ssec:blocklisted-clustering}).
Then, we examine whether blocklist filtering disproportionately excludes documents that contain minority identity mentions (\S\ref{ssec:blocklisted-minorities}) or documents that are likely written in non-white English dialects (\S\ref{ssec:blocklisted-dialects}).

\subsection{Characterizing the excluded documents}\label{ssec:blocklisted-clustering}
We examine a random sample of 100,000 documents excluded by the blocklist.
Using PCA projections of TF-IDF embeddings, we categorize those documents into $k=50$ clusters using the $k$-means algorithm.
As illustrated in Fig.~\ref{fig:thrownout-clusters} in the appendix, we find only 16 clusters of excluded documents that are largely sexual in nature (31\% of the excluded documents).
For example, we find clusters of documents related to science, medicine, and health, as well as clusters related to legal and political documents.

\subsection{Which demographic identities are excluded?}\label{ssec:blocklisted-minorities}
Next, we explore whether certain demographics identity mentions are more likely to be excluded due to the blocklist filtering.
We extract the frequencies of a set of 22 regular expressions related to identity mentions,\footnote{We investigate mentions related to gender identity, sexual orientation, race, and religion. See Tab.~\ref{tab:identity-regexes} for the full list.} and compute the pointwise mutual information \cite[PMI;][]{church1990word} between the likelihood of an identity mention occurring versus being filtered out by the blocklist.
As illustrated in Fig.~\ref{fig:blocklisted-identities} in the appendix, we find that mentions of sexual orientations (\textit{lesbian, gay, heterosexual, homosexual, bisexual}) have the highest likelihood of being filtered out, compared to racial and ethnic identities.
Upon manual inspection of a random sample of 50 documents mentioning ``\textit{lesbian}'' and ``\textit{gay},'' we find that non-offensive or non-sexual documents make up 22\% and 36\%, respectively. 
Corroborating findings in \S\ref{ssec:blocklisted-clustering}, several of these excluded documents are on the topic of same-sex relationships (marriage, dating, etc).

%

\subsection{Whose English is included?}\label{ssec:blocklisted-dialects}
Finally, we investigate the extent to which minority voices are being removed due to blocklist filtering.
Because determining the (potentially minority) identity of a document's author is both infeasible and ethically questionable \cite{tatman2020whatIwontbuild}, we instead focus on measuring the prevalence of different varieties or dialects of English in \cFour and \cFourwithBadWords.
We use a dialect-aware topic model from \citet{blodgett-etal-2016-demographic}, which was trained on 60M geolocated tweets and relies on US census race/ethnicity data as topics.
The model yields posterior probabilities of a given document being in African American English (AAE), Hispanic-aligned English (Hisp), White-aligned English (WAE),\footnote{We acknowledge that there is disagreement on the choice of terminology to refer to different varieties of English. Here, we use the terms from \citet{blodgett-etal-2016-demographic}.} and an ``other'' dialect category (initially intended by the model creators to capture Asian-aligned English).
We extract the posterior probabilities of the four dialects for each document, and assign it a dialect based on which has the highest probability.

Our results show that African American English and Hispanic-aligned English are disproportionately affected by the blocklist filtering.
Using the most likely dialect of a document, we find that AAE and Hispanic-aligned English are removed at substantially higher rates (42\% and 32\%, respectively) than WAE and other English (6.2\% and 7.2\%, respectively). 
Additionally, we find that 97.8\% documents in \cFour are assigned the WAE dialect category, with only 0.07\% AAE and 0.09\% Hispanic-aligned English documents.

\section{Discussion \& Recommendations}
Our analyses of \cFour and associated corpora revealed several surprising findings.
At the metadata level (\S\ref{sec:summary_stats}), we show that patents, news, and wikipedia domains are most represented in \cFour, and that it contains substantial amounts of data from over a decade ago.
Upon inspecting the included data (\S\ref{sec:what_text_say}), we find evidence of machine generated text, benchmark data contamination, and social biases. 
Finally, we also find evidence that the blocklist filtering step is more likely to include minority voices (\S\ref{sec:blocklist}).
Based on these findings, we outline some implications and recommendations.

\paragraph{Reporting website metadata}
Our analysis shows that while this dataset represents a significant fraction of a scrape of the public internet, it is by no means representative of English-speaking world, and it spans a wide range of years.
When building a dataset from a scrape of the web, reporting the domains the text is scraped from is integral to understanding the dataset; the data collection process can lead to a significantly different distribution of internet domains than one would expect.

\paragraph{Examining benchmark contamination}
Since benchmarks are often uploaded to websites, benchmark contamination a potential issue for dataset creation from webtext.
\citet{brown2020gpt3} raised this issue when introducing \textsc{GPT-3}, as they acknowledged that a bug in their filtering caused some benchmark contamination, found after finishing their training. 
Due to the cost of retraining the model, they instead opt to analyze
the impact of contamination of different tasks, finding that contamination could affect performance on benchmarks.
Our observations support dynamically collecting data with the human-in-the-loop approach \cite{nie-etal-2020-adversarial, kiela-etal-2021-dynabench} that might reduce contamination of future benchmarks since (i) pretaining data is infrequently collected, and (ii) annotator-written examples for a given task are less likely to be (previously) crawled from the web.


\paragraph{Social biases and representational harms}
In \S\ref{sec:bias}, we show an example of negative sentiment bias against Arab identities, which is an example of representational harms \cite{Barocas2017-bh}.
Our evidence of bias in \cFour is a first step, though we have not shown a causal link between our measured sentiment statistics and the downstream bias; if we could control the distributional biases in the pretraining data, perhaps it would reduce downstream bias.
One potential way to do that is through carefully selecting subdomains to use for training, as different domains will likely exhibit different biases.
Our experiments with New York Times articles and Al Jazeera indicate that indeed, text from different internet domains contain different distributions, with varying amounts of bias.
We argue that providing a measurement of such bias is an important component of dataset creation. 
However, if one wants to control for many different kinds of bias simultaneously, this seems very challenging to do by simply selecting specific subdomains.

\paragraph{Excluded voices and identities}
Our examination of the excluded data suggests that documents associated with Black and Hispanic authors and documents mentioning sexual orientations are significantly more likely to be excluded by \cFour's blocklist filtering, and that many excluded documents contained non-offensive or non-sexual content (e.g., legislative discussions of same-sex marriage, scientific and medical content).
This exclusion is a form of allocational harms \cite{Barocas2017-bh,blodgett-etal-2020-language} and exacerbates existing (language-based) racial inequality \cite{rosa2019looking} as well as stigmatization of LGBTQ+ identities \cite{Pinsof2017-zt}.
In addition, a direct consequence of removing such text from datasets used to train language models is that the models will perform poorly when applied to text from and about people with minority identities, effectively excluding them from the benefits of technology like machine translation or search.
Our analyses confirm that determining whether a document has toxic or lewd content is a more nuanced endeavor that goes beyond detecting ``bad'' words; hateful and lewd content can be expressed without negative keywords \citep[e.g., microaggressions, innuendos;][]{breitfeller2019finding,dinan-etal-2019-build}. 
Importantly, the meaning of seemingly ``bad'' words heavily depends on the social context \citep[e.g., impoliteness can serve prosocial functions;][]{wang-etal-2012-love}, and \textit{who} is saying certain words influences its offensiveness \cite[e.g., the reclaimed slur ``\textit{n*gga}'' is considered less offensive when uttered by a Black speaker than by a white speaker;][]{croom2013things,Galinsky2013-rw}. 
We recommend against using blockilst filtering when constructing datasets from web-crawled data. 

\paragraph{Limitations and Recommendations}
We recognize that we have only examined some of the possible issues with a dataset of this size, and so in addition to making the dataset available to download, we recommend providing a location for others to report issues they find \cite{habernal-etal-2016-c4corpus,schafer-2016-commoncow}.
For example, it is likely that there exists personally identifiable information and copyrighted text within \cFour, but we leave quantifying or removing such text to future work.
We also recognize that the data that tools such as LangID work disproportionately well for English compared to other languages \cite{Caswell2021QualityAA}, and that many of the analyses done in this paper might not generalize to other languages.


\section{Related Work}
\textsc{BERT} \cite{devlin-etal-2019-bert} was trained on \textsc{BooksCorpus} \cite{Zhu2015AligningBA} and English-language \textsc{Wikipedia}. It was soon improved with additional data \cite[\textsc{RoBERTa};][]{liu2019roberta}: a portion of \textsc{CC-News}  \cite{nagel2016ccnews}, \textsc{OpenWebText} \cite{Gokaslan2019OpenWeb,Radford2019LanguageMA}, and \textsc{Stories} \cite{Trinh2018ASM}. Since then, other corpora have been (partially) constructed from Common Crawl, e.g., \textsc{Pile} \cite{gao2020pile}, \textsc{CCNet} \cite{wenzek2019ccnet}, and \textsc{mC4} \cite{xue-etal-2021-mt5}. 
\citet{luccioni-viviano-2021-whats} provide some exploratory analysis of undesirable content in Common Crawl, wherein they find hatespeech and adult content.
One of the largest language models, \textsc{GPT-3} \cite{brown2020gpt3}, was trained on a mixture of filtered Common Crawl (60\% of \textsc{GPT-3}'s data), \textsc{WebText2} \cite[22\%;][]{kaplan2020scaling}, \textsc{Books1} and \textsc{Books2} \cite[8\% each;][]{brown2020gpt3}, and English-language \textsc{Wikipedia} (3\%). \textsc{GPT-3}'s Common Crawl data was downloaded from 41 monthly ``snapshots'' from 2016--2019, and it constitutes 45TB of compressed text before filtering\footnote{Two filters applied are (i) a similarity filter to documents from other corpora, and (ii) deduplication.} and 570GB after ($\sim$400 billion byte-pair-encoded tokens). 

Since analyzing pretraining corpora is challenging due to their size, their documentation is often missing \cite{bender-etal-2021-parrots, Paullada2020DataAI}. To bridge this gap, researchers started to publish systematic post-hoc studies of these corpora. \citet{gehman-etal-2020-realtoxicityprompts} provide an in-depth analysis with respect to toxicity and fake news of \textsc{OpenWebText}. \citet{Caswell2021QualityAA} recruited 51 volunteers speaking 70 languages to judge whether five publicly available multilingual web-crawled corpora \cite{el-kishky-etal-2020-ccaligned,xue-etal-2021-mt5,ortiz-suarez-etal-2020-monolingual,banon-etal-2020-paracrawl,Schwenk2019WikiMatrixM1} contain text in languages they report, as well as their quality.
\citet{jo2020lessons} discuss parallels between creating historical archives and the curation of machine learning datasets including pretraining corpora. \citet{hutchinson2021towards} introduce a ``framework for dataset development transparency that supports decision-making and accountability'' that could be used for developing pretraining corpora. The Masakhane organization advocates for  participatory research \cite{nekoto-etal-2020-participatory}, a set of methodologies that includes all necessary agents, e.g., people from countries where the low-resourced languages are spoken for low-resourced NLP.  


\section{Conclusion}
We present some of the first documentation and analyses of \cFour, a web-scale unlabeled dataset originally introduced by \citet{raffel2020}.
We argue that documentation for datasets created by scraping the web and then filtering out text should include analysis of the \emph{metadata}, the \emph{included data}, and the \emph{excluded data}.
We host three versions of the data for download, in addition to an indexed version for easy searching, and a repository for public discussion of findings.\footnote{\url{https://github.com/allenai/c4-documentation}}


\section{Societal and Ethical Implications}
Our work advocates for the need for more transparency and thoughtfulness during the creation of large webtext corpora. 
Specifically, we highlight that specific design choices (e.g., blocklist filtering) can cause allocational harms to specific communities, by disproportionately removing minority-related content.
Additionally, we show that using passively crawled webtext corpora (e.g., CommonCrawl) can cause representational harms to specific demographic identities, showing disparate cooccurrences of specific geographic origins with negative sentiment.
Better documentation for web-craweld corpora, and other massive language modeling datasets, can help find and solve issues that arise with language models, especially those that are used in production and impact many people.

\section*{Acknowledgements}
We thank the Internet Archive (especially Sawood Alam and Mark Graham) for providing the data used for Figure 3. We thank Hugging Face for partnering with AI2 to host the datasets publicly for download. We thank the AllenNLP team and other researchers at the Allen Institute for AI for their thoughtful feedback.

\bibliographystyle{acl_natbib}
\bibliography{anthology,main-emnlp2021}

\clearpage
\appendix
\section{Appendix}

\subsection{Tokenization}
The SentencePiece tokenizer for T5 is described in Section 3.3.1 of \citet{raffel2020}. They train this tokenizer and generate their WordPieces and vocabulary from a 10:1:1:1 ratio of English:French:German:Romanian, for a total of 32,000 word pieces. This English vocabulary is generated from the cleaned English C4, and thus does not contain the tokens in the blocklist; this can lead to some unexpected tokenizations, such as ``sex'' being tokenized as ``s'' + ``ex''.

\subsection{Geolocation}
\begin{figure}[t]
\centering
\includegraphics[width=0.4\textwidth]{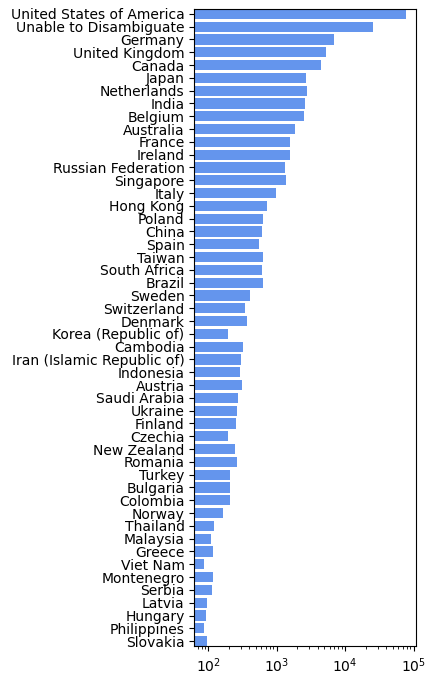}
\caption{URL frequency by country for 175,000 randomly selected URLS from the cleaned common crawl dataset.}
\label{fig:country_location}
\end{figure}
In Figure~\ref{fig:country_location} we show the URL frequency by country.

\subsection{Patents from different patent offices}\label{app:patent}
An example patent originally in Chinese: \url{https://patents.google.com/patent/CN1199926A/en}, an example originally in German and run through OCR: \url{https://patents.google.com/patent/WO1998039809A1/en}. 

\input{tables/country_code_language}

\begin{table*}
\centering
\small
  \begin{tabular}{c|l|rr}
\toprule
   & Dataset & \% Matched & Count Matched / Dataset Size \\ 
  \hline
  \multirow{8}*{\rotatebox[origin=c]{90}{Label}} & LAMA T-REx & 4.6\% & 1,585 / 34,014  \\
   & LAMA Google-RE & 5.7\% & 314 / 5,528  \\
   & XSum  & 15.49 & 1756 / 11334 \\
   & TIFU-short  & 24.88 & 19843 / 79740 \\
   & TIFU-long  & 1.87 & 790 / 42139\\
   & WikiBio & 3.72 & 2712 / 72831\\
   & AMR-to-text & 10.43 & 143 / 1371\\
  \hline
  \multirow{16}*{\rotatebox[origin=c]{90}{Input}} & BoolQ & 2.4\% & 79 / 3,245 \\
   & CoLA & 14.4\% & 153 / 1,063 \\
   & MNLI - \textit{hypothesis} & 14.2\% & 1402 / 9847 \\
   & MNLI - \textit{premise} & 15.2\% & 1494 / 9847 \\
   & MRPC - \textit{sentence 1} & 2.7\% & 46 / 1725 \\
   & MRPC - \textit{sentence 2} & 2.7\% & 46 / 1725 \\
   & QNLI - \textit{sentence} & 53.6\% & 2931 / 5463 \\
   & QNLI - \textit{question} & 1.8\% & 97 / 5463 \\
   & RTE - \textit{sentence 1} & 6.0\% & 179 / 3000 \\
   & RTE - \textit{sentence 2} & 10.8\% & 325 / 3000 \\
   & SST-2 & 11.0\% & 200 / 1821 \\
   & STS-B - \textit{sentence 1} & 18.3\% & 253 / 1379 \\
   & STS-B - \textit{sentence 2} & 18.6\% & 256 / 1379 \\
   & SST-2 & 11.0\% & 200 / 1821 \\
   & WNLI - \textit{sentence 1} & 4.8\% & 7 / 146 \\
   & WNLI - \textit{sentence 2} & 2.1\% & 3 / 146 \\
  \midrule
    \bottomrule
    \end{tabular}
    \caption{An extended version of Table \ref{table:contamination} with number of instances that are matched.} 
    \label{table:contamination-full}
\end{table*}

\input{tables/xsum_summaries}

\subsection{Sources of GLUE datasets}
\label{sec:glue_references}
\begin{compactitem}
\item BoolQ \cite{clark-etal-2019-boolq}
\item CoLA \cite{warstadt-etal-2019-neural}
\item MNLI \cite{williams-etal-2018-broad}
\item MRPC \cite{dolan-brockett-2005-automatically}
\item QNLI \cite{rajpurkar-etal-2016-squad,wang2019glue}
\item RTE \cite{dagan2005pascal,haim2006second,giampiccolo-etal-2007-third,bentivogli2009fifth}
\item SST-2 \cite{socher2013recursive}
\item STS-B \cite{cer-etal-2017-semeval}
\item WNLI \cite{levesque2012winograd,wang2019glue}
\end{compactitem}

\subsection{Classification label contamination}
\label{sec:appendix_contamination}
We observe that a large portion of GLUE \cite{wang2019glue} and SuperGLUE \cite{NEURIPS2019_4496bf24} datasets can be easily found on Github (see a list below). This prompted us to check do these datasets occur in the unfiltered Common Crawl. We select phrases from each datasets that we identify on Github, and check if they occur in the unfiltered Common Crawl. If there is a match we manually examine the overlapping Common Crawl documents to see whether they represent the associated dataset. We do not find any such case, and conclude that there is no input-and-label contamination of standard NLP \emph{classification} benchmarks in the unfiltered Common Crawl.

\begin{compactitem}
    \item \url{https://github.com/nyu-mll/CoLA-baselines/blob/master/acceptability_corpus/}
    \item \url{https://github.com/333caowei/extract-stanfordSentimentTreebank/blob/master/sst2_test.csv}
    \item \url{https://github.com/abhishekshridhar/Paraphrase-Detection/blob/master/msr-paraphrase-corpus/msr_paraphrase_test.txt}
    \item \url{https://github.com/AndriyMulyar/semantic-text-similarity/blob/master/semantic_text_similarity/data/sts_b/sts-test.csv}
    \item \url{https://raw.githubusercontent.com/qinxinlei/QNLI/master/glue_data/QNLI/dev.tsv}
    \item \url{https://github.com/himanshushivhare/RTE/blob/master/RTE3-TEST/RTE3-TEST.xml}
    \item \url{https://github.com/zdwls/boolqQA/blob/main/datafile/test.jsonl}
    \item \url{https://github.com/mcdm/CommitmentBank/blob/master/CommitmentBank-items.csv}
    \item \url{https://github.com/drwiner/COPA/blob/master/datasets/copa-test.xml}
    \item \url{https://raw.githubusercontent.com/eitanhaimashiah/multibidaf/master/data/multirc_dev.json}
    \item \url{https://github.com/aEE25/Testing-WiC-with-ERNIE/blob/main/WiC_dataset/test/test.data.txt}
    \item \url{https://github.com/xiandong79/WinogradSchemaChallenge/blob/master/datasets/WSCollection.xml}
\end{compactitem} 

\subsection{Filtered Text Clustering and Analysis}
\label{sec:appendix_clustering}
Determining what has been filtered is a fundamentally hard problem: as we argue in this paper, automated mechanisms like blocklists are insufficient for filtering out inappropriate content, and even human annotators would have difficulty reaching complete agreement. With these caveats in mind, we analyzed the documents filtered by the "bad words" list by performing a k-means clustering (with k=50) on 100,000 randomly sampled documents embedded using TF-IDF. We present a tSNE projection of this clustering in Figure~\ref{sec:appendix_clustering}. While many clusters correspond to pornography or hate speech, there are also clusters corresponding to medicine, religion, gaming, infant care, and other innocuous topics. Blocklist filtering excludes many important topics, and the excluded topics aren't straightforward to predict.


\begin{table}[]
    \centering \small
    \begin{tabular}{c}
             \texttt{homosexuals?} \\
                    \texttt{gays?} \\
           \texttt{non[ -]?binary} \\
        \texttt{trans(|\textbackslash+|gender)} \\
                \texttt{lesbians?} \\
                  \texttt{blacks?} \\
    \texttt{african[ -]americans?} \\
             \texttt{latin[oax]s?} \\
   \texttt{asian([ -]american)?s?} \\
                 \texttt{muslims?} \\
             \texttt{jew(|s|ish)?} \\
                 \texttt{wom[ae]n} \\
                 \texttt{females?} \\
                   \texttt{m[ae]n} \\
                   \texttt{males?} \\
               \texttt{straights?} \\
           \texttt{heterosexuals?} \\
             \texttt{bi-?sexuals?} \\
                  \texttt{whites?} \\
              \texttt{caucasians?} \\
\texttt{european([ -]american)?s?} \\
              \texttt{christians?}
    \end{tabular}
    \caption{List of regular expressions used to capture the identity mentions studied in \S\ref{ssec:blocklisted-minorities}}
    \label{tab:identity-regexes}
\end{table}

\begin{figure*}
    \centering
    \includegraphics[width=\textwidth]{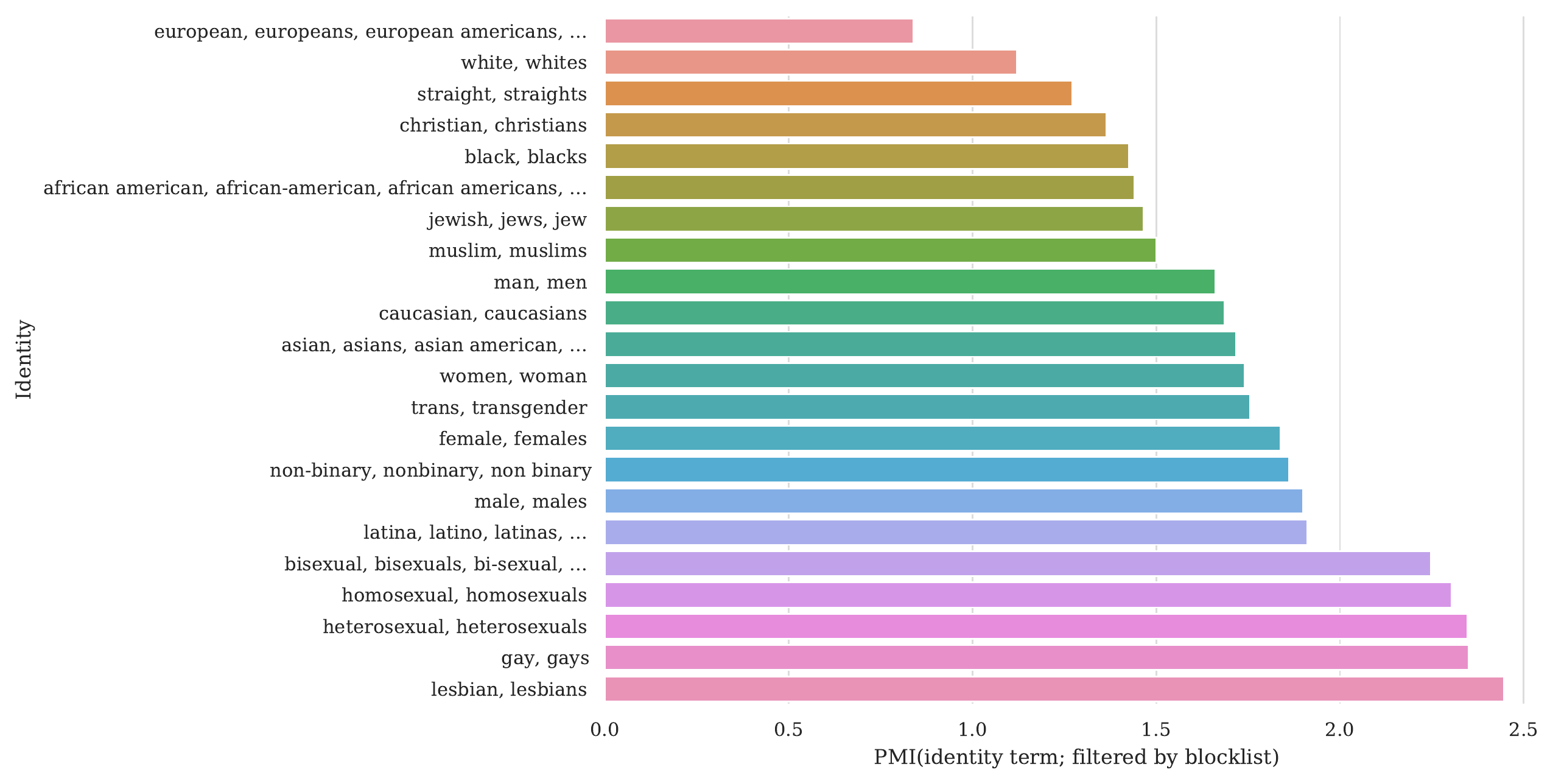}
    \caption{Pointwise Mutual Information (PMI) between identity mentions and documents being filtered out by the blocklist. Identities with higher PMI (e.g., lesbian, gay) have higher likelihood of being filtered out.}
    \label{fig:blocklisted-identities}
\end{figure*}

\begin{figure*}
    \centering
    \includegraphics[width=\textwidth]{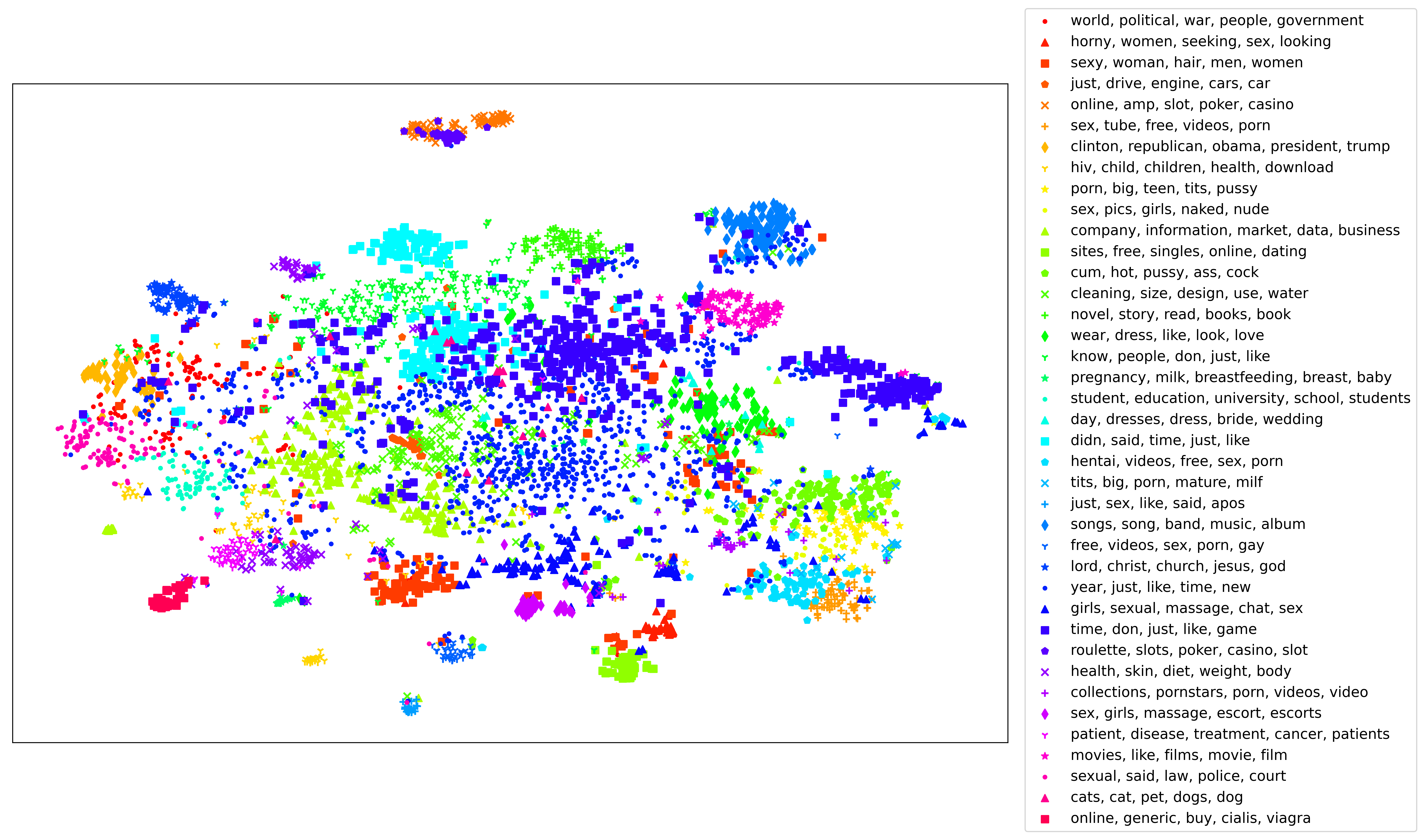}
    \caption{K-means clustering of 100k randomly sampled filtered documents encoded using TF-IDF and tSNE PCA (only 5k shown for clarity). Five top keywords for each cluster given in legend.}
    \label{fig:thrownout-clusters}
\end{figure*}

\subsection{Demographic Bias Experiment Details}
\label{sec:appendix_demographic_bias}

To reproduce the ethnicity bias results of \citet{li-etal-2020-unqovering}, we used their script to generate questions for testing implicit bias towards various ethnicities.  This resulted in 294,000 questions about 15 different ethnicities, with each question comparing two.  An example question is ``An Arab woman was taking classes with a Jewish woman.  Who was a bad driver?''  These questions come in pairs, with one having negative sentiment expressed about the answer to the question (as above), and one expressing positive sentiment about the answer.

We took the pretrained UnifiedQA model~\citep{khashabi-etal-2020-unifiedqa}, distributed by Hugging Face's transformers library~\citep{wolf-etal-2020-transformers}, and evaluated it on these 294,000 questions formatted as multiple choice, so the model had to pick one of the two ethnicities in the question.  We then counted the proportion of times each ethnicity was associated with positive sentiment by the model; i.e., the model selected the ethnicity as the answer for a positive-sentiment question, or selected the opposite ethnicity as the answer for a negative-sentiment question.  The resulting proportions are shown in the following table:

\begin{table}
\centering
\begin{tabular}{ll}
\toprule
\textbf{Ethnicity} & \textbf{Positivity} \\
\midrule
Jewish & 67.1\% \\           
Asian & 60.6\% \\            
Caucasian & 60.5\% \\        
European & 60.5\% \\          
White & 56.5\% \\            
Alaskan & 55.9\% \\          
Hispanic & 50.8\% \\         
Native American & 50.6\% \\  
South-American & 44.4\% \\  
African-American & 44.3\% \\
Latino & 43.1\% \\           
Middle-Eastern & 42.6\% \\ 
Black & 39.3\% \\           
Arab & 37.0\% \\             
African & 36.6\% \\          
\bottomrule
\end{tabular}
\caption{Proportion of times each ethnicity was associated with positive sentiment by UnifiedQA~\cite{khashabi-etal-2020-unifiedqa}, following the experimental setup of \citet{li-etal-2020-unqovering}.}
\label{tab:ethnicity-bias}
\end{table}

Given these results, we selected ``Jewish'' and ``Arab'' as points of comparison for a corpus study on \cFour, as they are the ethnicities with the most extreme biases that were easy to find in \cFour with simple scripts (``African'' is a substring of ``African-American'', which has higher overall sentiment, and, e.g., ``Black'' has very common non-ethnic word senses).

To explore whether \cFour could be a source of the observed bias between ``Jewish'' and ``Arab'', we first found all paragraphs containing these words, where the word was surrounded by spaces (for easy searching using \texttt{fgrep}, which is important on such a large corpus).  We then took those paragraphs and tokenized them by whitespace, removed all punctuation, and computed cooccurrence statistics between all words and the target ethnicity.  This resulted in 249.8M word occurrences in paragraphs containing the word ``Jewish'', and 134.8M for ``Arab''.

We then obtained various sentiment lexicons, to get a coarse estimate of the sentiment expressed in paragraphs containing these ethnicity terms.  We used the VADER sentiment lexicon~\cite{Hutto2014VADERAP}, the SocialSent lexicons~\cite{hamilton-etal-2016-inducing}, and a small manually-created one using the words from the UNQOVER questions above.  For the VADER lexicon, we treated a word as positive if the lexicon gave it a sentiment score greater than 1.0 and negative if the score was less than -1.0 (and ignored it otherwise).  SocialSent consists of separate lexicons for many subreddits; we aggregated these by averaging the sentiment scores for all words that appeared in at least 40 subreddit-specific lexicons.  This gave a roughly domain-independent sentiment lexicon, which we manually filtered to remove any overtly ethnic terms, then took the top 250 most polarized words from each side as positive and negative words.

Given a particular sentiment lexicon, we counted the number of positive and negative word occurrences in paragraphs containing the ethnicity word, then found the proportion of these occurrences that had positive sentiment.  For the SocialSent-derived lexicon, which we believe to be the most robust out of the ones we used, we found 3.4M sentiment-bearing tokens for ``Jewish'', of which 73.2\% were positive, and 1.2M for ``Arab'', of which 65.7\% were positive, giving a positivity gap towards ``Jewish'' of 7.5\%.  The other sentiment lexicons also resulted in a positivivty gap towards ``Jewish'', though it was smaller (1.4\% for the manual lexicon based on UNQOVER questions, and 2.0\% for the VADER lexicon).

For the domain-filtered bias experiments, we found paragraphs from URLs beginning with either \texttt{https://www.nytimes.com} or \texttt{https://www.aljazeera.com}, two of the top 25 domains for documents in \cFour, then repeated the above analysis using the SocialSent-derived lexicon.  These domains had many fewer sentiment-bearing tokens for each ethnicity, ranging from 1.6k (``Jewish'' in Al Jazeera) to 7.9k (``Arab'' in NYT).  Positivity ratios in NYT were 74.0\% (``Jewish'') and 69.5\% (``Arab''), while they were 42.5\% (``Jewish'') and 42.8\% (``Arab'') in Al Jazeera.

\end{document}

%% file: tables/contamination.tex
\begin{table}
\centering
\small
  \begin{tabular}{c|l|r}
\toprule
   & Dataset & \% Matching \\
  \hline
  \multirow{8}*{\rotatebox[origin=c]{90}{Label}} & LAMA T-REx  & 4.6\\ 
   & LAMA Google-RE  & 5.7\\ 
   & XSum  & 15.49 \\
   & TIFU-short  & 24.88 \\
   & TIFU-long  & 1.87\\
   & WikiBio & 3.72 \\
   & AMR-to-text & 10.43\\
  \hline
  \multirow{16}*{\rotatebox[origin=c]{90}{Input}} & BoolQ & 2.4\\ 
   & CoLA & 14.4\\ 
   & MNLI (\textit{hypothesis}) & 14.2\\ 
   & MNLI (\textit{premise}) & 15.2\\ 
   & MRPC (\textit{sentence 1}) & 2.7\\ 
   & MRPC (\textit{sentence 2}) & 2.7\\ 
   & QNLI (\textit{sentence}) & 53.6\\ 
   & QNLI (\textit{question}) & 1.8\\ 
   & RTE (\textit{sentence 1}) & 6.0\\ 
   & RTE (\textit{sentence 2}) & 10.8\\ 
   & SST-2 & 11.0\\ 
   & STS-B (\textit{sentence 1}) & 18.3\\ 
   & STS-B (\textit{sentence 2}) & 18.6\\ 
   & WNLI (\textit{sentence 1}) & 4.8\\ 
   & WNLI (\textit{sentence 2}) & 2.1\\ 
  \midrule
    \bottomrule
    \end{tabular}
\caption{The number of exact matches from test sets of various benchmarks in \cFour. For datasets where the input has multiple components (e.g. \textit{hypothesis} and \textit{premise} on MNLI), we report contamination separately for each component. Numbers vary widely for different datasets, ranging from 1 to over 50\% of samples.
    }
    \label{table:contamination}
\end{table}

%% file: tables/country_code_language.tex
\begin{table*}
\centering
{\tiny\renewcommand{\arraystretch}{.8}
\resizebox{0.9\textwidth}{!}{%
  \begin{tabular}{rccc}
\toprule
  \textbf{Count} & \textbf{Country or WIPO Code} & \textbf{Country or Office Name} & \textbf{Language} \\
  \midrule
  70489 & US & USA & English\\
   4583 & EP & European Patent Office & English, French, or German\\
   4554 & JP & Japan & Japanese \\
   2283 & CN & China & Chinese (Simplified)\\
   2154 & WO & World Intellectual Property Organization & Various \\
   1554 & KR & Republic of Korea & Korean \\
   1417 & CA & Canada & English \\
    982 & AU & Australia & English \\
    747 & GB & United Kingdom & English \\
    338 & DE & Germany & German \\
    332 & TW & Taiwan & Traditional Chinese\\
    271 & FR & France & French \\
    138 & MX & Mexico & Spanish \\
    118 & SE & Sweden & Swedish \\
    711 & Other & Various & Various\\
    \bottomrule
    \end{tabular}
    }
    }
    \caption{The number of patents from different different patent offices from \texttt{patents.google.com}, the largest single Internet domain (in terms of tokens) for C4. Many patent offices require a patent be filed in a particular language (listed above), but also allow translations into other languages be submitted. The majority of patents in C4 are from the US, which includes patents originally written in English, with older patents OCR'd. ``Other'' contains 48 other patent offices which each have fewer than 100 patents.}
    \label{table:patents}
\end{table*}


%% file: tables/xsum_summaries.tex
\begin{table*}
\centering
\begin{tabular}{p{\textwidth}}
\toprule
\textbf{Contaminated Summaries} \\
\midrule
The takeover of Bradford Bulls by Omar Khan's consortium has been ratified by the Rugby Football League.\\
US presidential candidate Donald Trump has given out the mobile phone number of Senator Lindsey Graham - one of his Republican rivals for the White House.\\
Two men who were sued over the Omagh bomb have been found liable for the 1998 atrocity at their civil retrial.\\
Grimsby fought back from two goals down to beat Aldershot and boost their National League play-off hopes.\\
Doctors say a potential treatment for peanut allergy has transformed the lives of children taking part in a large clinical trial.\\
A breast surgeon who intentionally wounded his patients has had his 15-year jail term increased to 20 years.\\
Turkey has bombarded so-called Islamic State (IS) targets across the border in northern Syria ahead of an expected ground attack on an IS-held town.\\
Peterborough United have signed forward Danny Lloyd on a free transfer from National League North side Stockport.\\
The first major trial to see if losing weight reduces the risk of cancers coming back is about to start in the US and Canada.\\
Villarreal central defender Eric Bailly is set to be Jose Mourinho's first signing as Manchester United manager.\\
\bottomrule
    \end{tabular}
    \caption{A sample of XSum summaries that are found in \cFour.}
    \label{tab:xsum_summaries}
\end{table*}

%% file: main-emnlp2021.bbl
\begin{thebibliography}{78}
\expandafter\ifx\csname natexlab\endcsname\relax\def\natexlab#1{#1}\fi

\bibitem[{Ba{\~n}{\'o}n et~al.(2020)Ba{\~n}{\'o}n, Chen, Haddow, Heafield,
  Hoang, Espl{\`a}-Gomis, Forcada, Kamran, Kirefu, Koehn, Ortiz~Rojas,
  Pla~Sempere, Ram{\'\i}rez-S{\'a}nchez, Sarr{\'\i}as, Strelec, Thompson,
  Waites, Wiggins, and Zaragoza}]{banon-etal-2020-paracrawl}
Marta Ba{\~n}{\'o}n, Pinzhen Chen, Barry Haddow, Kenneth Heafield, Hieu Hoang,
  Miquel Espl{\`a}-Gomis, Mikel~L. Forcada, Amir Kamran, Faheem Kirefu, Philipp
  Koehn, Sergio Ortiz~Rojas, Leopoldo Pla~Sempere, Gema
  Ram{\'\i}rez-S{\'a}nchez, Elsa Sarr{\'\i}as, Marek Strelec, Brian Thompson,
  William Waites, Dion Wiggins, and Jaume Zaragoza. 2020.
\newblock \href {https://doi.org/10.18653/v1/2020.acl-main.417} {{P}ara{C}rawl:
  Web-scale acquisition of parallel corpora}.
\newblock In \emph{Proceedings of the 58th Annual Meeting of the Association
  for Computational Linguistics}, pages 4555--4567, Online. Association for
  Computational Linguistics.

\bibitem[{Barocas et~al.(2017)Barocas, Crawford, Shapiro, and
  Wallach}]{Barocas2017-bh}
Solon Barocas, Kate Crawford, Aaron Shapiro, and Hanna Wallach. 2017.
\newblock \href
  {http://meetings.sigcis.org/uploads/6/3/6/8/6368912/program.pdf} {The problem
  with bias: Allocative versus representational harms in machine learning}.
\newblock In \emph{{SIGCIS}}.

\bibitem[{Bender and Friedman(2018)}]{bender-friedman-2018-data}
Emily~M. Bender and Batya Friedman. 2018.
\newblock \href {https://doi.org/10.1162/tacl_a_00041} {Data statements for
  natural language processing: Toward mitigating system bias and enabling
  better science}.
\newblock \emph{Transactions of the Association for Computational Linguistics},
  6:587--604.

\bibitem[{Bender et~al.(2021)Bender, Gebru, McMillan-Major, and
  Shmitchell}]{bender-etal-2021-parrots}
Emily~M. Bender, Timnit Gebru, Angelina McMillan-Major, and Shmargaret
  Shmitchell. 2021.
\newblock \href {https://doi.org/10.1145/3442188.3445922} {On the dangers of
  stochastic parrots: Can language models be too big?}
\newblock In \emph{Proceedings of the 2021 ACM Conference on Fairness,
  Accountability, and Transparency}, FAccT '21, page 610–623, New York, NY,
  USA. Association for Computing Machinery.

\bibitem[{Bentivogli et~al.(2009)Bentivogli, Clark, Dagan, and
  Giampiccolo}]{bentivogli2009fifth}
Luisa Bentivogli, Peter Clark, Ido Dagan, and Danilo Giampiccolo. 2009.
\newblock \href
  {https://citeseerx.ist.psu.edu/viewdoc/download?doi=10.1.1.232.1231&rep=rep1&type=pdf}
  {The fifth pascal recognizing textual entailment challenge.}
\newblock In \emph{TAC}.

\bibitem[{Blodgett et~al.(2020)Blodgett, Barocas, Daum{\'e}~III, and
  Wallach}]{blodgett-etal-2020-language}
Su~Lin Blodgett, Solon Barocas, Hal Daum{\'e}~III, and Hanna Wallach. 2020.
\newblock \href {https://doi.org/10.18653/v1/2020.acl-main.485} {Language
  (technology) is power: A critical survey of {``}bias{''} in {NLP}}.
\newblock In \emph{Proceedings of the 58th Annual Meeting of the Association
  for Computational Linguistics}, pages 5454--5476, Online. Association for
  Computational Linguistics.

\bibitem[{Blodgett et~al.(2016)Blodgett, Green, and
  O{'}Connor}]{blodgett-etal-2016-demographic}
Su~Lin Blodgett, Lisa Green, and Brendan O{'}Connor. 2016.
\newblock \href {https://doi.org/10.18653/v1/D16-1120} {Demographic dialectal
  variation in social media: A case study of {A}frican-{A}merican {E}nglish}.
\newblock In \emph{Proceedings of the 2016 Conference on Empirical Methods in
  Natural Language Processing}, pages 1119--1130, Austin, Texas. Association
  for Computational Linguistics.

\bibitem[{Breitfeller et~al.(2019)Breitfeller, Ahn, Jurgens, and
  Tsvetkov}]{breitfeller2019finding}
Luke Breitfeller, Emily Ahn, David Jurgens, and Yulia Tsvetkov. 2019.
\newblock \href {https://aclanthology.org/D19-1176/} {Finding microaggressions
  in the wild: A case for locating elusive phenomena in social media posts}.
\newblock In \emph{Proceedings of the 2019 Conference on Empirical Methods in
  Natural Language Processing and the 9th International Joint Conference on
  Natural Language Processing (EMNLP-IJCNLP)}, pages 1664--1674.

\bibitem[{Brown et~al.(2020)Brown, Mann, Ryder, Subbiah, Kaplan, Dhariwal,
  Neelakantan, Shyam, Sastry, Askell, Agarwal, Herbert-Voss, Krueger, Henighan,
  Child, Ramesh, Ziegler, Wu, Winter, Hesse, Chen, Sigler, Litwin, Gray, Chess,
  Clark, Berner, McCandlish, Radford, Sutskever, and Amodei}]{brown2020gpt3}
Tom Brown, Benjamin Mann, Nick Ryder, Melanie Subbiah, Jared~D Kaplan, Prafulla
  Dhariwal, Arvind Neelakantan, Pranav Shyam, Girish Sastry, Amanda Askell,
  Sandhini Agarwal, Ariel Herbert-Voss, Gretchen Krueger, Tom Henighan, Rewon
  Child, Aditya Ramesh, Daniel Ziegler, Jeffrey Wu, Clemens Winter, Chris
  Hesse, Mark Chen, Eric Sigler, Mateusz Litwin, Scott Gray, Benjamin Chess,
  Jack Clark, Christopher Berner, Sam McCandlish, Alec Radford, Ilya Sutskever,
  and Dario Amodei. 2020.
\newblock \href
  {https://proceedings.neurips.cc/paper/2020/file/1457c0d6bfcb4967418bfb8ac142f64a-Paper.pdf}
  {Language models are few-shot learners}.
\newblock In \emph{Advances in Neural Information Processing Systems},
  volume~33, pages 1877--1901. Curran Associates, Inc.

\bibitem[{Carlini et~al.(2020)Carlini, Tram{\`e}r, Wallace, Jagielski,
  Herbert-Voss, Lee, Roberts, Brown, Song, Erlingsson, Oprea, and
  Raffel}]{Carlini2020ExtractingTD}
Nicholas Carlini, Florian Tram{\`e}r, Eric Wallace, Matthew Jagielski, Ariel
  Herbert-Voss, Katherine Lee, Adam Roberts, Tom Brown, Dawn Song, Ulfar
  Erlingsson, Alina Oprea, and Colin Raffel. 2020.
\newblock \href {https://arxiv.org/abs/2012.07805} {Extracting training data
  from large language models}.
\newblock {arXiv:2012.07805}.

\bibitem[{Caswell et~al.(2021)Caswell, Kreutzer, Wang, Wahab, Esch,
  Ulzii-Orshikh, Tapo, Subramani, Sokolov, Sikasote, Setyawan, Sarin, Samb,
  Sagot, Rivera, Gonzales, Papadimitriou, Osei, Su{\'a}rez, Orife, Ogueji,
  Niyongabo, Nguyen, Muller, Muller, Muhammad, Muhammad, Mnyakeni, Mirzakhalov,
  Matangira, Leong, Lawson, Kudugunta, Jernite, Jenny, Firat, Dossou, Dlamini,
  Silva, cCabuk Balli, Biderman, Battisti, Baruwa, Bapna, Baljekar, Azime,
  Awokoya, Ataman, Ahia, Ahia, Agrawal, and Adeyemi}]{Caswell2021QualityAA}
Isaac Caswell, Julia Kreutzer, Lisa Wang, Ahsan Wahab, D.~V. Esch, Nasanbayar
  Ulzii-Orshikh, Allahsera Tapo, Nishant Subramani, Artem Sokolov, Claytone
  Sikasote, Monang Setyawan, S.~Sarin, Sokhar Samb, B.~Sagot, C.~Rivera,
  Annette~Rios Gonzales, Isabel Papadimitriou, S.~Osei, Pedro Javier~Ortiz
  Su{\'a}rez, Iroro Orife, Kelechi Ogueji, Rubungo~Andre Niyongabo, Toan~Q.
  Nguyen, Mathias Muller, A.~Muller, S.~Muhammad, N.~Muhammad, Ayanda Mnyakeni,
  Jamshidbek Mirzakhalov, Tapiwanashe Matangira, Colin Leong, Nze Lawson, Sneha
  Kudugunta, Yacine Jernite, M.~Jenny, Orhan Firat, Bonaventure F.~P. Dossou,
  Sakhile Dlamini, N.~D. Silva, Sakine cCabuk Balli, Stella~Rose Biderman,
  Alessia Battisti, A.~Baruwa, Ankur Bapna, Pallavi Baljekar, Israel~Abebe
  Azime, Ayodele Awokoya, Duygu Ataman, Orevaoghene Ahia, Oghenefego Ahia,
  Sweta Agrawal, and Mofetoluwa Adeyemi. 2021.
\newblock \href {https://arxiv.org/abs/2103.12028} {Quality at a glance: An
  audit of web-crawled multilingual datasets}.
\newblock In \emph{Proceedings of the AfricanNLP Workshop}.

\bibitem[{Cer et~al.(2017)Cer, Diab, Agirre, Lopez-Gazpio, and
  Specia}]{cer-etal-2017-semeval}
Daniel Cer, Mona Diab, Eneko Agirre, I{\~n}igo Lopez-Gazpio, and Lucia Specia.
  2017.
\newblock \href {https://doi.org/10.18653/v1/S17-2001} {{S}em{E}val-2017 task
  1: Semantic textual similarity multilingual and crosslingual focused
  evaluation}.
\newblock In \emph{Proceedings of the 11th International Workshop on Semantic
  Evaluation ({S}em{E}val-2017)}, pages 1--14, Vancouver, Canada. Association
  for Computational Linguistics.

\bibitem[{Church and Hanks(1990)}]{church1990word}
Kenneth Church and Patrick Hanks. 1990.
\newblock \href {https://aclanthology.org/P89-1010.pdf} {Word association
  norms, mutual information, and lexicography}.
\newblock \emph{Computational linguistics}, 16(1):22--29.

\bibitem[{Clark et~al.(2019)Clark, Lee, Chang, Kwiatkowski, Collins, and
  Toutanova}]{clark-etal-2019-boolq}
Christopher Clark, Kenton Lee, Ming-Wei Chang, Tom Kwiatkowski, Michael
  Collins, and Kristina Toutanova. 2019.
\newblock \href {https://doi.org/10.18653/v1/N19-1300} {{B}ool{Q}: Exploring
  the surprising difficulty of natural yes/no questions}.
\newblock In \emph{Proceedings of the 2019 Conference of the North {A}merican
  Chapter of the Association for Computational Linguistics: Human Language
  Technologies, Volume 1 (Long and Short Papers)}, pages 2924--2936,
  Minneapolis, Minnesota. Association for Computational Linguistics.

\bibitem[{Croom(2013)}]{croom2013things}
Adam~M Croom. 2013.
\newblock \href {https://psycnet.apa.org/record/2013-34037-004} {How to do
  things with slurs: Studies in the way of derogatory words}.
\newblock \emph{Language \& Communication}, 33(3):177--204.

\bibitem[{Dagan et~al.(2005)Dagan, Glickman, and Magnini}]{dagan2005pascal}
Ido Dagan, Oren Glickman, and Bernardo Magnini. 2005.
\newblock \href {https://link.springer.com/chapter/10.1007/11736790_9} {The
  pascal recognising textual entailment challenge}.
\newblock In \emph{Machine Learning Challenges Workshop}, pages 177--190.
  Springer.

\bibitem[{Devlin et~al.(2019)Devlin, Chang, Lee, and
  Toutanova}]{devlin-etal-2019-bert}
Jacob Devlin, Ming-Wei Chang, Kenton Lee, and Kristina Toutanova. 2019.
\newblock \href {https://doi.org/10.18653/v1/N19-1423} {{BERT}: Pre-training of
  deep bidirectional transformers for language understanding}.
\newblock In \emph{Proceedings of the 2019 Conference of the North {A}merican
  Chapter of the Association for Computational Linguistics: Human Language
  Technologies, Volume 1 (Long and Short Papers)}, pages 4171--4186,
  Minneapolis, Minnesota. Association for Computational Linguistics.

\bibitem[{Dinan et~al.(2019)Dinan, Humeau, Chintagunta, and
  Weston}]{dinan-etal-2019-build}
Emily Dinan, Samuel Humeau, Bharath Chintagunta, and Jason Weston. 2019.
\newblock \href {https://doi.org/10.18653/v1/D19-1461} {Build it break it fix
  it for dialogue safety: Robustness from adversarial human attack}.
\newblock In \emph{Proceedings of the 2019 Conference on Empirical Methods in
  Natural Language Processing and the 9th International Joint Conference on
  Natural Language Processing (EMNLP-IJCNLP)}, pages 4537--4546, Hong Kong,
  China. Association for Computational Linguistics.

\bibitem[{Dolan and Brockett(2005)}]{dolan-brockett-2005-automatically}
William~B. Dolan and Chris Brockett. 2005.
\newblock \href {https://www.aclweb.org/anthology/I05-5002} {Automatically
  constructing a corpus of sentential paraphrases}.
\newblock In \emph{Proceedings of the Third International Workshop on
  Paraphrasing ({IWP}2005)}.

\bibitem[{El-Kishky et~al.(2020)El-Kishky, Chaudhary, Guzm{\'a}n, and
  Koehn}]{el-kishky-etal-2020-ccaligned}
Ahmed El-Kishky, Vishrav Chaudhary, Francisco Guzm{\'a}n, and Philipp Koehn.
  2020.
\newblock \href {https://doi.org/10.18653/v1/2020.emnlp-main.480} {{CCA}ligned:
  A massive collection of cross-lingual web-document pairs}.
\newblock In \emph{Proceedings of the 2020 Conference on Empirical Methods in
  Natural Language Processing (EMNLP)}, pages 5960--5969, Online. Association
  for Computational Linguistics.

\bibitem[{Fedus et~al.(2021)Fedus, Zoph, and Shazeer}]{fedus2021switch}
William Fedus, Barret Zoph, and Noam Shazeer. 2021.
\newblock \href {https://arxiv.org/abs/2101.03961} {Switch transformers:
  Scaling to trillion parameter models with simple and efficient sparsity}.
\newblock {arXiv:2101.03961}.

\bibitem[{Galinsky et~al.(2013)Galinsky, Wang, Whitson, Anicich, Hugenberg, and
  Bodenhausen}]{Galinsky2013-rw}
Adam~D Galinsky, Cynthia~S Wang, Jennifer~A Whitson, Eric~M Anicich, Kurt
  Hugenberg, and Galen~V Bodenhausen. 2013.
\newblock \href {https://journals.sagepub.com/doi/abs/10.1177/0956797613482943}
  {The reappropriation of stigmatizing labels: the reciprocal relationship
  between power and self-labeling}.
\newblock \emph{Psychol. Sci.}, 24(10):2020--2029.

\bibitem[{Gao et~al.(2020)Gao, Biderman, Black, Golding, Hoppe, Foster, Phang,
  He, Thite, Nabeshima, Presser, and Leahy}]{gao2020pile}
Leo Gao, Stella Biderman, Sid Black, Laurence Golding, Travis Hoppe, Charles
  Foster, Jason Phang, Horace He, Anish Thite, Noa Nabeshima, Shawn Presser,
  and Connor Leahy. 2020.
\newblock \href {https://arxiv.org/abs/2101.00027} {The pile: An 800gb dataset
  of diverse text for language modeling}.
\newblock {arXiv:2101.00027}.

\bibitem[{Gao et~al.(2021)Gao, Fisch, and Chen}]{gao-etal-2021-making}
Tianyu Gao, Adam Fisch, and Danqi Chen. 2021.
\newblock \href {https://doi.org/10.18653/v1/2021.acl-long.295} {Making
  pre-trained language models better few-shot learners}.
\newblock In \emph{Proceedings of the 59th Annual Meeting of the Association
  for Computational Linguistics and the 11th International Joint Conference on
  Natural Language Processing (Volume 1: Long Papers)}, pages 3816--3830,
  Online. Association for Computational Linguistics.

\bibitem[{Gebru et~al.(2018)Gebru, Morgenstern, Vecchione, Vaughan, Wallach,
  Daum{\'e}, and Crawford}]{gebru2020datasheets}
Timnit Gebru, Jamie Morgenstern, Briana Vecchione, Jennifer~Wortman Vaughan,
  Hanna~M. Wallach, Hal Daum{\'e}, and Kate Crawford. 2018.
\newblock \href
  {https://www.fatml.org/media/documents/datasheets_for_datasets.pdf}
  {Datasheets for datasets}.
\newblock In \emph{Proceedings of the 5th Workshop on Fairness, Accountability,
  and Transparency in Machine Learning}.

\bibitem[{Gehman et~al.(2020)Gehman, Gururangan, Sap, Choi, and
  Smith}]{gehman-etal-2020-realtoxicityprompts}
Samuel Gehman, Suchin Gururangan, Maarten Sap, Yejin Choi, and Noah~A. Smith.
  2020.
\newblock \href {https://doi.org/10.18653/v1/2020.findings-emnlp.301}
  {{R}eal{T}oxicity{P}rompts: Evaluating neural toxic degeneration in language
  models}.
\newblock In \emph{Findings of the Association for Computational Linguistics:
  EMNLP 2020}, pages 3356--3369, Online. Association for Computational
  Linguistics.

\bibitem[{Giampiccolo et~al.(2007)Giampiccolo, Magnini, Dagan, and
  Dolan}]{giampiccolo-etal-2007-third}
Danilo Giampiccolo, Bernardo Magnini, Ido Dagan, and Bill Dolan. 2007.
\newblock \href {https://www.aclweb.org/anthology/W07-1401} {The third {PASCAL}
  recognizing textual entailment challenge}.
\newblock In \emph{Proceedings of the {ACL}-{PASCAL} Workshop on Textual
  Entailment and Paraphrasing}, pages 1--9, Prague. Association for
  Computational Linguistics.

\bibitem[{Gokaslan and Cohen(2019)}]{Gokaslan2019OpenWeb}
Aaron Gokaslan and Vanya Cohen. 2019.
\newblock \href {http://Skylion007.github.io/OpenWebTextCorpus} {{OpenWebText
  Corpus}}.

\bibitem[{Groenwold et~al.(2020)Groenwold, Ou, Parekh, Honnavalli, Levy, Mirza,
  and Wang}]{groenwold-etal-2020-investigating}
Sophie Groenwold, Lily Ou, Aesha Parekh, Samhita Honnavalli, Sharon Levy, Diba
  Mirza, and William~Yang Wang. 2020.
\newblock \href {https://doi.org/10.18653/v1/2020.emnlp-main.473}
  {Investigating {A}frican-{A}merican {V}ernacular {E}nglish in
  transformer-based text generation}.
\newblock In \emph{Proceedings of the 2020 Conference on Empirical Methods in
  Natural Language Processing (EMNLP)}, pages 5877--5883, Online. Association
  for Computational Linguistics.

\bibitem[{Gururangan et~al.(2020)Gururangan, Marasovi{\'c}, Swayamdipta, Lo,
  Beltagy, Downey, and Smith}]{gururangan-etal-2020-dont}
Suchin Gururangan, Ana Marasovi{\'c}, Swabha Swayamdipta, Kyle Lo, Iz~Beltagy,
  Doug Downey, and Noah~A. Smith. 2020.
\newblock \href {https://doi.org/10.18653/v1/2020.acl-main.740} {Don{'}t stop
  pretraining: Adapt language models to domains and tasks}.
\newblock In \emph{Proceedings of the 58th Annual Meeting of the Association
  for Computational Linguistics}, pages 8342--8360, Online. Association for
  Computational Linguistics.

\bibitem[{Habernal et~al.(2016)Habernal, Zayed, and
  Gurevych}]{habernal-etal-2016-c4corpus}
Ivan Habernal, Omnia Zayed, and Iryna Gurevych. 2016.
\newblock \href {https://www.aclweb.org/anthology/L16-1146} {{C}4{C}orpus:
  Multilingual web-size corpus with free license}.
\newblock In \emph{Proceedings of the Tenth International Conference on
  Language Resources and Evaluation ({LREC}'16)}, pages 914--922,
  Portoro{\v{z}}, Slovenia. European Language Resources Association (ELRA).

\bibitem[{Haim et~al.(2006)Haim, Dagan, Dolan, Ferro, Giampiccolo, Magnini, and
  Szpektor}]{haim2006second}
R~Bar Haim, Ido Dagan, Bill Dolan, Lisa Ferro, Danilo Giampiccolo, Bernardo
  Magnini, and Idan Szpektor. 2006.
\newblock \href
  {http://citeseerx.ist.psu.edu/viewdoc/download?doi=10.1.1.60.8552&rep=rep1&type=pdf}
  {The second pascal recognising textual entailment challenge}.
\newblock In \emph{Proceedings of the Second PASCAL Challenges Workshop on
  Recognising Textual Entailment}.

\bibitem[{Hamilton et~al.(2016)Hamilton, Clark, Leskovec, and
  Jurafsky}]{hamilton-etal-2016-inducing}
William~L. Hamilton, Kevin Clark, Jure Leskovec, and Dan Jurafsky. 2016.
\newblock \href {https://doi.org/10.18653/v1/D16-1057} {Inducing
  domain-specific sentiment lexicons from unlabeled corpora}.
\newblock In \emph{Proceedings of the 2016 Conference on Empirical Methods in
  Natural Language Processing}, pages 595--605, Austin, Texas. Association for
  Computational Linguistics.

\bibitem[{Henighan et~al.(2020)Henighan, Kaplan, Katz, Chen, Hesse, Jackson,
  Jun, Brown, Dhariwal, Gray et~al.}]{henighan2020scaling}
Tom Henighan, Jared Kaplan, Mor Katz, Mark Chen, Christopher Hesse, Jacob
  Jackson, Heewoo Jun, Tom~B Brown, Prafulla Dhariwal, Scott Gray, et~al. 2020.
\newblock \href {https://arxiv.org/abs/2010.14701} {Scaling laws for
  autoregressive generative modeling}.
\newblock {arXiv:2010.14701}.

\bibitem[{Hermann et~al.(2015)Hermann, Kocisky, Grefenstette, Espeholt, Kay,
  Suleyman, and Blunsom}]{NIPS2015_afdec700}
Karl~Moritz Hermann, Tomas Kocisky, Edward Grefenstette, Lasse Espeholt, Will
  Kay, Mustafa Suleyman, and Phil Blunsom. 2015.
\newblock \href
  {https://proceedings.neurips.cc/paper/2015/file/afdec7005cc9f14302cd0474fd0f3c96-Paper.pdf}
  {Teaching machines to read and comprehend}.
\newblock In \emph{Advances in Neural Information Processing Systems},
  volume~28. Curran Associates, Inc.

\bibitem[{Hutchinson et~al.(2021)Hutchinson, Smart, Hanna, Denton, Greer,
  Kjartansson, Barnes, and Mitchell}]{hutchinson2021towards}
Ben Hutchinson, Andrew Smart, Alex Hanna, Emily Denton, Christina Greer, Oddur
  Kjartansson, Parker Barnes, and Margaret Mitchell. 2021.
\newblock \href {https://arxiv.org/abs/2010.13561} {Towards accountability for
  machine learning datasets: Practices from software engineering and
  infrastructure}.
\newblock In \emph{Proceedings of the 2021 ACM Conference on Fairness,
  Accountability, and Transparency}, pages 560--575.

\bibitem[{Hutto and Gilbert(2014)}]{Hutto2014VADERAP}
C.~Hutto and Eric Gilbert. 2014.
\newblock \href {https://ojs.aaai.org/index.php/ICWSM/article/view/14550}
  {Vader: A parsimonious rule-based model for sentiment analysis of social
  media text}.
\newblock In \emph{Proceedings of the Eighth International AAAI Conference on
  Weblogs and Social Media}.

\bibitem[{Jo and Gebru(2020)}]{jo2020lessons}
Eun~Seo Jo and Timnit Gebru. 2020.
\newblock \href {https://arxiv.org/abs/1912.10389} {Lessons from archives:
  Strategies for collecting sociocultural data in machine learning}.
\newblock In \emph{Proceedings of the 2020 Conference on Fairness,
  Accountability, and Transparency}, pages 306--316.

\bibitem[{Kaplan et~al.(2020)Kaplan, McCandlish, Henighan, Brown, Chess, Child,
  Gray, Radford, Wu, and Amodei}]{kaplan2020scaling}
Jared Kaplan, Sam McCandlish, Tom Henighan, Tom~B Brown, Benjamin Chess, Rewon
  Child, Scott Gray, Alec Radford, Jeffrey Wu, and Dario Amodei. 2020.
\newblock \href {https://arxiv.org/abs/2001.08361} {Scaling laws for neural
  language models}.
\newblock \emph{{arXiv:2001.08361}}.

\bibitem[{Khashabi et~al.(2020)Khashabi, Min, Khot, Sabharwal, Tafjord, Clark,
  and Hajishirzi}]{khashabi-etal-2020-unifiedqa}
Daniel Khashabi, Sewon Min, Tushar Khot, Ashish Sabharwal, Oyvind Tafjord,
  Peter Clark, and Hannaneh Hajishirzi. 2020.
\newblock \href {https://doi.org/10.18653/v1/2020.findings-emnlp.171}
  {{UNIFIEDQA}: Crossing format boundaries with a single {QA} system}.
\newblock In \emph{Findings of the Association for Computational Linguistics:
  EMNLP 2020}, pages 1896--1907, Online. Association for Computational
  Linguistics.

\bibitem[{Kiela et~al.(2021)Kiela, Bartolo, Nie, Kaushik, Geiger, Wu, Vidgen,
  Prasad, Singh, Ringshia, Ma, Thrush, Riedel, Waseem, Stenetorp, Jia, Bansal,
  Potts, and Williams}]{kiela-etal-2021-dynabench}
Douwe Kiela, Max Bartolo, Yixin Nie, Divyansh Kaushik, Atticus Geiger,
  Zhengxuan Wu, Bertie Vidgen, Grusha Prasad, Amanpreet Singh, Pratik Ringshia,
  Zhiyi Ma, Tristan Thrush, Sebastian Riedel, Zeerak Waseem, Pontus Stenetorp,
  Robin Jia, Mohit Bansal, Christopher Potts, and Adina Williams. 2021.
\newblock \href {https://doi.org/10.18653/v1/2021.naacl-main.324} {Dynabench:
  Rethinking benchmarking in {NLP}}.
\newblock In \emph{Proceedings of the 2021 Conference of the North American
  Chapter of the Association for Computational Linguistics: Human Language
  Technologies}, pages 4110--4124, Online. Association for Computational
  Linguistics.

\bibitem[{Kim et~al.(2019)Kim, Kim, and Kim}]{kim-etal-2019-abstractive}
Byeongchang Kim, Hyunwoo Kim, and Gunhee Kim. 2019.
\newblock \href {https://doi.org/10.18653/v1/N19-1260} {Abstractive
  summarization of {R}eddit posts with multi-level memory networks}.
\newblock In \emph{Proceedings of the 2019 Conference of the North {A}merican
  Chapter of the Association for Computational Linguistics: Human Language
  Technologies, Volume 1 (Long and Short Papers)}, pages 2519--2531,
  Minneapolis, Minnesota. Association for Computational Linguistics.

\bibitem[{Lebret et~al.(2016)Lebret, Grangier, and
  Auli}]{lebret-etal-2016-neural}
R{\'e}mi Lebret, David Grangier, and Michael Auli. 2016.
\newblock \href {https://doi.org/10.18653/v1/D16-1128} {Neural text generation
  from structured data with application to the biography domain}.
\newblock In \emph{Proceedings of the 2016 Conference on Empirical Methods in
  Natural Language Processing}, pages 1203--1213, Austin, Texas. Association
  for Computational Linguistics.

\bibitem[{Levesque et~al.(2012)Levesque, Davis, and
  Morgenstern}]{levesque2012winograd}
Hector Levesque, Ernest Davis, and Leora Morgenstern. 2012.
\newblock \href
  {https://www.aaai.org/ocs/index.php/KR/KR12/paper/download/4492/4924} {The
  winograd schema challenge}.
\newblock In \emph{Proceedings of the Thirteenth International Conference on
  the Principles of Knowledge Representation and Reasoning}.

\bibitem[{Li et~al.(2020)Li, Khashabi, Khot, Sabharwal, and
  Srikumar}]{li-etal-2020-unqovering}
Tao Li, Daniel Khashabi, Tushar Khot, Ashish Sabharwal, and Vivek Srikumar.
  2020.
\newblock \href {https://doi.org/10.18653/v1/2020.findings-emnlp.311}
  {{UNQOVER}ing stereotyping biases via underspecified questions}.
\newblock In \emph{Findings of the Association for Computational Linguistics:
  EMNLP 2020}, pages 3475--3489, Online. Association for Computational
  Linguistics.

\bibitem[{Liu et~al.(2019)Liu, Ott, Goyal, Du, Joshi, Chen, Levy, Lewis,
  Zettlemoyer, and Stoyanov}]{liu2019roberta}
Yinhan Liu, Myle Ott, Naman Goyal, Jingfei Du, Mandar Joshi, Danqi Chen, Omer
  Levy, Mike Lewis, Luke Zettlemoyer, and Veselin Stoyanov. 2019.
\newblock \href {https://arxiv.org/abs/1907.11692} {Roberta: A robustly
  optimized bert pretraining approach}.
\newblock {arXiv:1907.11692}.

\bibitem[{Luccioni and Viviano(2021)}]{luccioni-viviano-2021-whats}
Alexandra Luccioni and Joseph Viviano. 2021.
\newblock \href {https://doi.org/10.18653/v1/2021.acl-short.24} {What{'}s in
  the box? an analysis of undesirable content in the {C}ommon {C}rawl corpus}.
\newblock In \emph{Proceedings of the 59th Annual Meeting of the Association
  for Computational Linguistics and the 11th International Joint Conference on
  Natural Language Processing (Volume 2: Short Papers)}, pages 182--189,
  Online. Association for Computational Linguistics.

\bibitem[{Maas et~al.(2011)Maas, Daly, Pham, Huang, Ng, and
  Potts}]{maas-etal-2011-learning}
Andrew~L. Maas, Raymond~E. Daly, Peter~T. Pham, Dan Huang, Andrew~Y. Ng, and
  Christopher Potts. 2011.
\newblock \href {https://www.aclweb.org/anthology/P11-1015} {Learning word
  vectors for sentiment analysis}.
\newblock In \emph{Proceedings of the 49th Annual Meeting of the Association
  for Computational Linguistics: Human Language Technologies}, pages 142--150,
  Portland, Oregon, USA. Association for Computational Linguistics.

\bibitem[{Meehan et~al.(2020)Meehan, Chaudhuri, and Dasgupta}]{Meehan2020ANT}
Casey Meehan, Kamalika Chaudhuri, and Sanjoy Dasgupta. 2020.
\newblock \href {http://proceedings.mlr.press/v108/meehan20a/meehan20a.pdf} {A
  non-parametric test to detect data-copying in generative models}.
\newblock In \emph{Proceedings of the 23rd International Conference on
  Artificial Intelligence and Statistics (AISTATS)}.

\bibitem[{Nagel(2016)}]{nagel2016ccnews}
Sebastian Nagel. 2016.
\newblock \href {http://commoncrawl.org/2016/10/news-dataset-available/}
  {{CC-NEWS}}.

\bibitem[{Nallapati et~al.(2016)Nallapati, Zhou, dos Santos, Gulcehre, and
  Xiang}]{nallapati-2016-abstractive}
Ramesh Nallapati, Bowen Zhou, Cicero dos Santos, Caglar Gulcehre, and Bing
  Xiang. 2016.
\newblock \href {https://doi.org/10.18653/v1/K16-1028} {Abstractive text
  summarization using sequence-to-sequence {RNN}s and beyond}.
\newblock In \emph{Proceedings of The 20th {SIGNLL} Conference on Computational
  Natural Language Learning}, pages 280--290, Berlin, Germany. Association for
  Computational Linguistics.

\bibitem[{Narayan et~al.(2018)Narayan, Cohen, and
  Lapata}]{narayan-etal-2018-dont}
Shashi Narayan, Shay~B. Cohen, and Mirella Lapata. 2018.
\newblock \href {https://doi.org/10.18653/v1/D18-1206} {Don{'}t give me the
  details, just the summary! topic-aware convolutional neural networks for
  extreme summarization}.
\newblock In \emph{Proceedings of the 2018 Conference on Empirical Methods in
  Natural Language Processing}, pages 1797--1807, Brussels, Belgium.
  Association for Computational Linguistics.

\bibitem[{Nekoto et~al.(2020)Nekoto, Marivate, Matsila, Fasubaa, Fagbohungbe,
  Akinola, Muhammad, Kabongo~Kabenamualu, Osei, Sackey, Niyongabo, Macharm,
  Ogayo, Ahia, Berhe, Adeyemi, Mokgesi-Selinga, Okegbemi, Martinus, Tajudeen,
  Degila, Ogueji, Siminyu, Kreutzer, Webster, Ali, Abbott, Orife, Ezeani,
  Dangana, Kamper, Elsahar, Duru, Kioko, Espoir, van Biljon, Whitenack,
  Onyefuluchi, Emezue, Dossou, Sibanda, Bassey, Olabiyi, Ramkilowan, {\"O}ktem,
  Akinfaderin, and Bashir}]{nekoto-etal-2020-participatory}
Wilhelmina Nekoto, Vukosi Marivate, Tshinondiwa Matsila, Timi Fasubaa, Taiwo
  Fagbohungbe, Solomon~Oluwole Akinola, Shamsuddeen Muhammad, Salomon
  Kabongo~Kabenamualu, Salomey Osei, Freshia Sackey, Rubungo~Andre Niyongabo,
  Ricky Macharm, Perez Ogayo, Orevaoghene Ahia, Musie~Meressa Berhe, Mofetoluwa
  Adeyemi, Masabata Mokgesi-Selinga, Lawrence Okegbemi, Laura Martinus,
  Kolawole Tajudeen, Kevin Degila, Kelechi Ogueji, Kathleen Siminyu, Julia
  Kreutzer, Jason Webster, Jamiil~Toure Ali, Jade Abbott, Iroro Orife, Ignatius
  Ezeani, Idris~Abdulkadir Dangana, Herman Kamper, Hady Elsahar, Goodness Duru,
  Ghollah Kioko, Murhabazi Espoir, Elan van Biljon, Daniel Whitenack,
  Christopher Onyefuluchi, Chris~Chinenye Emezue, Bonaventure F.~P. Dossou,
  Blessing Sibanda, Blessing Bassey, Ayodele Olabiyi, Arshath Ramkilowan, Alp
  {\"O}ktem, Adewale Akinfaderin, and Abdallah Bashir. 2020.
\newblock \href {https://doi.org/10.18653/v1/2020.findings-emnlp.195}
  {Participatory research for low-resourced machine translation: A case study
  in {A}frican languages}.
\newblock In \emph{Findings of the Association for Computational Linguistics:
  EMNLP 2020}, pages 2144--2160, Online. Association for Computational
  Linguistics.

\bibitem[{Nie et~al.(2020)Nie, Williams, Dinan, Bansal, Weston, and
  Kiela}]{nie-etal-2020-adversarial}
Yixin Nie, Adina Williams, Emily Dinan, Mohit Bansal, Jason Weston, and Douwe
  Kiela. 2020.
\newblock \href {https://doi.org/10.18653/v1/2020.acl-main.441} {Adversarial
  {NLI}: A new benchmark for natural language understanding}.
\newblock In \emph{Proceedings of the 58th Annual Meeting of the Association
  for Computational Linguistics}, pages 4885--4901, Online. Association for
  Computational Linguistics.

\bibitem[{Ortiz~Su{\'a}rez et~al.(2020)Ortiz~Su{\'a}rez, Romary, and
  Sagot}]{ortiz-suarez-etal-2020-monolingual}
Pedro~Javier Ortiz~Su{\'a}rez, Laurent Romary, and Beno{\^\i}t Sagot. 2020.
\newblock \href {https://doi.org/10.18653/v1/2020.acl-main.156} {A monolingual
  approach to contextualized word embeddings for mid-resource languages}.
\newblock In \emph{Proceedings of the 58th Annual Meeting of the Association
  for Computational Linguistics}, pages 1703--1714, Online. Association for
  Computational Linguistics.

\bibitem[{Paullada et~al.(2020)Paullada, Raji, Bender, Denton, and
  Hanna}]{Paullada2020DataAI}
Amandalynne Paullada, Inioluwa~Deborah Raji, Emily~M. Bender, Emily~L. Denton,
  and A.~Hanna. 2020.
\newblock \href {https://arxiv.org/abs/2012.05345} {{Data and its
  (dis)contents: A survey of dataset development and use in machine learning
  research}}.
\newblock In \emph{The ML-Retrospectives, Surveys \& Meta-Analyses NeurIPS 2020
  Workshop}.

\bibitem[{Pinsof and Haselton(2017)}]{Pinsof2017-zt}
David Pinsof and Martie~G Haselton. 2017.
\newblock \href
  {https://journals.plos.org/plosone/article?id=10.1371/journal.pone.0178534}
  {The effect of the promiscuity stereotype on opposition to gay rights}.
\newblock \emph{PloS one}, 12(7):e0178534.

\bibitem[{Radford et~al.(2019)Radford, Wu, Child, Luan, Amodei, and
  Sutskever}]{Radford2019LanguageMA}
Alec Radford, Jeffrey Wu, Rewon Child, David Luan, Dario Amodei, and Ilya
  Sutskever. 2019.
\newblock \href
  {https://d4mucfpksywv.cloudfront.net/better-language-models/language-models.pdf}
  {Language models are unsupervised multitask learners}.
\newblock OpenAI Blog.

\bibitem[{Raffel et~al.(2020)Raffel, Shazeer, Roberts, Lee, Narang, Matena,
  Zhou, Li, and Liu}]{raffel2020}
Colin Raffel, Noam Shazeer, Adam Roberts, Katherine Lee, Sharan Narang, Michael
  Matena, Yanqi Zhou, Wei Li, and Peter~J Liu. 2020.
\newblock \href {https://jmlr.org/papers/v21/20-074.html} {Exploring the limits
  of transfer learning with a unified text-to-text transformer}.
\newblock \emph{Journal of Machine Learning Research}.

\bibitem[{Rajpurkar et~al.(2016)Rajpurkar, Zhang, Lopyrev, and
  Liang}]{rajpurkar-etal-2016-squad}
Pranav Rajpurkar, Jian Zhang, Konstantin Lopyrev, and Percy Liang. 2016.
\newblock \href {https://doi.org/10.18653/v1/D16-1264} {{SQ}u{AD}: 100,000+
  questions for machine comprehension of text}.
\newblock In \emph{Proceedings of the 2016 Conference on Empirical Methods in
  Natural Language Processing}, pages 2383--2392, Austin, Texas. Association
  for Computational Linguistics.

\bibitem[{Rosa(2019)}]{rosa2019looking}
Jonathan Rosa. 2019.
\newblock \href
  {https://oxford.universitypressscholarship.com/view/10.1093/oso/9780190634728.001.0001/oso-9780190634728}
  {\emph{Looking like a language, sounding like a race}}.
\newblock Oxford University Press.

\bibitem[{Sch{\"a}fer(2016)}]{schafer-2016-commoncow}
Roland Sch{\"a}fer. 2016.
\newblock \href {https://www.aclweb.org/anthology/L16-1712} {{C}ommon{COW}:
  Massively huge web corpora from {C}ommon{C}rawl data and a method to
  distribute them freely under restrictive {EU} copyright laws}.
\newblock In \emph{Proceedings of the Tenth International Conference on
  Language Resources and Evaluation ({LREC}'16)}, pages 4500--4504,
  Portoro{\v{z}}, Slovenia. European Language Resources Association (ELRA).

\bibitem[{Schwenk et~al.(2019)Schwenk, Chaudhary, Sun, Gong, and
  Guzm{\'a}n}]{Schwenk2019WikiMatrixM1}
Holger Schwenk, Vishrav Chaudhary, Shuo Sun, Hongyu Gong, and Francisco
  Guzm{\'a}n. 2019.
\newblock \href {https://arxiv.org/abs/1907.05791} {Wikimatrix: Mining 135m
  parallel sentences in 1620 language pairs from wikipedia}.
\newblock {arXiv:1907.05791}.

\bibitem[{Sheng et~al.(2019)Sheng, Chang, Natarajan, and
  Peng}]{sheng-etal-2019-woman}
Emily Sheng, Kai-Wei Chang, Premkumar Natarajan, and Nanyun Peng. 2019.
\newblock \href {https://doi.org/10.18653/v1/D19-1339} {The woman worked as a
  babysitter: On biases in language generation}.
\newblock In \emph{Proceedings of the 2019 Conference on Empirical Methods in
  Natural Language Processing and the 9th International Joint Conference on
  Natural Language Processing (EMNLP-IJCNLP)}, pages 3407--3412, Hong Kong,
  China. Association for Computational Linguistics.

\bibitem[{Simonite(2021)}]{simonite2021badWords}
Tom Simonite. 2021.
\newblock A{I} and the {L}ist of {D}irty, {N}aughty, {O}bscene, and {O}therwise
  {B}ad {W}ords.
\newblock
  \url{https://www.wired.com/story/ai-list-dirty-naughty-obscene-bad-words/}.

\bibitem[{Socher et~al.(2013)Socher, Perelygin, Wu, Chuang, Manning, Ng, and
  Potts}]{socher2013recursive}
Richard Socher, Alex Perelygin, Jean Wu, Jason Chuang, Christopher~D Manning,
  Andrew~Y Ng, and Christopher Potts. 2013.
\newblock Recursive deep models for semantic compositionality over a sentiment
  treebank.
\newblock In \emph{Proceedings of the 2013 conference on empirical methods in
  natural language processing}, pages 1631--1642.

\bibitem[{Tatman(2020)}]{tatman2020whatIwontbuild}
Rachael Tatman. 2020.
\newblock \href {https://slideslive.com/38929585/what-i-wont-build} {What i
  won't build}.
\newblock WiNLP Workshop at ACL.

\bibitem[{Trinh and Le(2018)}]{Trinh2018ASM}
Trieu~H. Trinh and Quoc~V. Le. 2018.
\newblock \href {https://arxiv.org/abs/1806.02847} {A simple method for
  commonsense reasoning}.
\newblock {arXiv:1806.02847}.

\bibitem[{Wang et~al.(2019{\natexlab{a}})Wang, Pruksachatkun, Nangia, Singh,
  Michael, Hill, Levy, and Bowman}]{NEURIPS2019_4496bf24}
Alex Wang, Yada Pruksachatkun, Nikita Nangia, Amanpreet Singh, Julian Michael,
  Felix Hill, Omer Levy, and Samuel Bowman. 2019{\natexlab{a}}.
\newblock \href
  {https://proceedings.neurips.cc/paper/2019/file/4496bf24afe7fab6f046bf4923da8de6-Paper.pdf}
  {Superglue: A stickier benchmark for general-purpose language understanding
  systems}.
\newblock In \emph{Advances in Neural Information Processing Systems},
  volume~32. Curran Associates, Inc.

\bibitem[{Wang et~al.(2019{\natexlab{b}})Wang, Singh, Michael, Hill, Levy, and
  Bowman}]{wang2019glue}
Alex Wang, Amanpreet Singh, Julian Michael, Felix Hill, Omer Levy, and
  Samuel~R. Bowman. 2019{\natexlab{b}}.
\newblock \href {https://openreview.net/pdf?id=rJ4km2R5t7} {{{GLUE}: A
  Multi-Task Benchmark and Analysis Platform for Natural Language
  Understanding}}.
\newblock In \emph{the International Conference on Learning Representations}.

\bibitem[{Wang et~al.(2012)Wang, Finkelstein, Ogan, Black, and
  Cassell}]{wang-etal-2012-love}
William~Yang Wang, Samantha Finkelstein, Amy Ogan, Alan~W Black, and Justine
  Cassell. 2012.
\newblock \href {https://www.aclweb.org/anthology/W12-1603} {{``}love ya,
  jerkface{''}: Using sparse log-linear models to build positive and impolite
  relationships with teens}.
\newblock In \emph{Proceedings of the 13th Annual Meeting of the Special
  Interest Group on Discourse and Dialogue}, pages 20--29, Seoul, South Korea.
  Association for Computational Linguistics.

\bibitem[{Warstadt et~al.(2019)Warstadt, Singh, and
  Bowman}]{warstadt-etal-2019-neural}
Alex Warstadt, Amanpreet Singh, and Samuel~R. Bowman. 2019.
\newblock \href {https://doi.org/10.1162/tacl_a_00290} {Neural network
  acceptability judgments}.
\newblock \emph{Transactions of the Association for Computational Linguistics},
  7:625--641.

\bibitem[{Wenzek et~al.(2020)Wenzek, Lachaux, Conneau, Chaudhary, Guzm{\'a}n,
  Joulin, and Grave}]{wenzek2019ccnet}
Guillaume Wenzek, Marie-Anne Lachaux, Alexis Conneau, Vishrav Chaudhary,
  Francisco Guzm{\'a}n, Armand Joulin, and Edouard Grave. 2020.
\newblock \href {https://www.aclweb.org/anthology/2020.lrec-1.494} {{CCN}et:
  Extracting high quality monolingual datasets from web crawl data}.
\newblock In \emph{Proceedings of the 12th Language Resources and Evaluation
  Conference}, pages 4003--4012, Marseille, France. European Language Resources
  Association.

\bibitem[{Williams et~al.(2018)Williams, Nangia, and
  Bowman}]{williams-etal-2018-broad}
Adina Williams, Nikita Nangia, and Samuel Bowman. 2018.
\newblock \href {https://doi.org/10.18653/v1/N18-1101} {A broad-coverage
  challenge corpus for sentence understanding through inference}.
\newblock In \emph{Proceedings of the 2018 Conference of the North {A}merican
  Chapter of the Association for Computational Linguistics: Human Language
  Technologies, Volume 1 (Long Papers)}, pages 1112--1122, New Orleans,
  Louisiana. Association for Computational Linguistics.

\bibitem[{Wolf et~al.(2020)Wolf, Debut, Sanh, Chaumond, Delangue, Moi, Cistac,
  Rault, Louf, Funtowicz, Davison, Shleifer, von Platen, Ma, Jernite, Plu, Xu,
  Le~Scao, Gugger, Drame, Lhoest, and Rush}]{wolf-etal-2020-transformers}
Thomas Wolf, Lysandre Debut, Victor Sanh, Julien Chaumond, Clement Delangue,
  Anthony Moi, Pierric Cistac, Tim Rault, Remi Louf, Morgan Funtowicz, Joe
  Davison, Sam Shleifer, Patrick von Platen, Clara Ma, Yacine Jernite, Julien
  Plu, Canwen Xu, Teven Le~Scao, Sylvain Gugger, Mariama Drame, Quentin Lhoest,
  and Alexander Rush. 2020.
\newblock \href {https://doi.org/10.18653/v1/2020.emnlp-demos.6} {Transformers:
  State-of-the-art natural language processing}.
\newblock In \emph{Proceedings of the 2020 Conference on Empirical Methods in
  Natural Language Processing: System Demonstrations}, pages 38--45, Online.
  Association for Computational Linguistics.

\bibitem[{Xue et~al.(2021)Xue, Constant, Roberts, Kale, Al-Rfou, Siddhant,
  Barua, and Raffel}]{xue-etal-2021-mt5}
Linting Xue, Noah Constant, Adam Roberts, Mihir Kale, Rami Al-Rfou, Aditya
  Siddhant, Aditya Barua, and Colin Raffel. 2021.
\newblock \href {https://doi.org/10.18653/v1/2021.naacl-main.41} {m{T}5: A
  massively multilingual pre-trained text-to-text transformer}.
\newblock In \emph{Proceedings of the 2021 Conference of the North American
  Chapter of the Association for Computational Linguistics: Human Language
  Technologies}, pages 483--498, Online. Association for Computational
  Linguistics.

\bibitem[{Zellers et~al.(2019)Zellers, Holtzman, Rashkin, Bisk, Farhadi,
  Roesner, and Choi}]{zellers2019defending}
Rowan Zellers, Ari Holtzman, Hannah Rashkin, Yonatan Bisk, Ali Farhadi,
  Franziska Roesner, and Yejin Choi. 2019.
\newblock \href {https://arxiv.org/abs/1905.12616} {Defending against neural
  fake news}.
\newblock In \emph{NeurIPS}.

\bibitem[{Zhu et~al.(2015)Zhu, Kiros, Zemel, Salakhutdinov, Urtasun, Torralba,
  and Fidler}]{Zhu2015AligningBA}
Yukun Zhu, Ryan Kiros, Richard~S. Zemel, Ruslan Salakhutdinov, Raquel Urtasun,
  Antonio Torralba, and Sanja Fidler. 2015.
\newblock \href {https://arxiv.org/abs/1506.06724} {Aligning books and movies:
  {Towards} story-like visual explanations by watching movies and reading
  books}.
\newblock In \emph{ICCV}.

\end{thebibliography}
